\documentclass[letterpaper]{article} 
\usepackage{aaai24}  
\usepackage{times}  
\usepackage{helvet}  
\usepackage{courier}  
\usepackage[hyphens]{url}  
\usepackage{graphicx} 
\usepackage{natbib}  
\usepackage{caption} 
\usepackage{algorithm}
\usepackage{algorithmic}

\usepackage{newfloat}
\usepackage{listings}
\DeclareCaptionStyle{ruled}{labelfont=normalfont,labelsep=colon,strut=off} 
\floatstyle{ruled}
\newfloat{listing}{tb}{lst}{}
\floatname{listing}{Listing}
\usepackage{bibentry}

\usepackage{amsmath,amsfonts,bm}

\newcommand{\estq}[1]{\overline{#1}^{\pi}}

\def\eqref#1{equation~\ref{#1}}

\def\1{\bm{1}}

\DeclareMathAlphabet{\mathsfit}{\encodingdefault}{\sfdefault}{m}{sl}
\SetMathAlphabet{\mathsfit}{bold}{\encodingdefault}{\sfdefault}{bx}{n}

\newcommand{\E}{\mathbb{E}}

\newcommand{\parab}[1]{\vspace{0.05in}\noindent\textbf{#1}}

\usepackage{enumitem}
\usepackage{pbox}
\usepackage{multirow}
\usepackage{tabularx}
\usepackage{makecell}
\usepackage{mathtools}
\usepackage{booktabs}
\usepackage{xcolor}
\usepackage{xspace}
\usepackage[labelformat=empty, font=normalsize,labelfont=normalsize]{subcaption}

\makeatletter
\newcommand{\labeltext}[2]{%
  \@bsphack
  \csname phantomsection\endcsname 
  \def\@currentlabel{#1}{\label{#2}}%
  \@esphack
}
\makeatother

\title{CrystalBox: Future-Based Explanations for Input-Driven Deep RL Systems}
\author {
    Sagar Patel\textsuperscript{\rm 1},
    Sangeetha Abdu Jyothi\textsuperscript{\rm 1, \rm 2},
    Nina Narodytska\textsuperscript{\rm 2}
}
\affiliations {
    \textsuperscript{\rm 1}University of California, Irvine\\
    \textsuperscript{\rm 2}VMware Research\\
    sagar.patel@uci.edu, sangeetha.aj@uci.edu, nina.narodytska@broadcom.com
}

\newcommand{\comment}[1]{}
\newcommand{\fmname}{CrystalBox\xspace}

\begin{document}

\maketitle

\begin{abstract}

We present CrystalBox, a novel, model-agnostic, posthoc explainability framework for Deep Reinforcement Learning (DRL) controllers in the large family of input-driven environments which includes computer systems. We combine the natural decomposability of reward functions in input-driven environments with the explanatory power of decomposed returns. We propose an efficient algorithm to generate future-based explanations across both discrete and continuous control environments. Using applications such as adaptive bitrate streaming and congestion control, we demonstrate CrystalBox's capability to generate high-fidelity explanations. We further illustrate its higher utility across three practical use cases: contrastive explanations, network observability, and guided reward design, as opposed to prior explainability techniques that identify salient features.

\end{abstract} 

\begin{figure*}[t]
\centering
\begin{subfigure}{.9\textwidth}
    \centering
	\includegraphics[width=\linewidth]{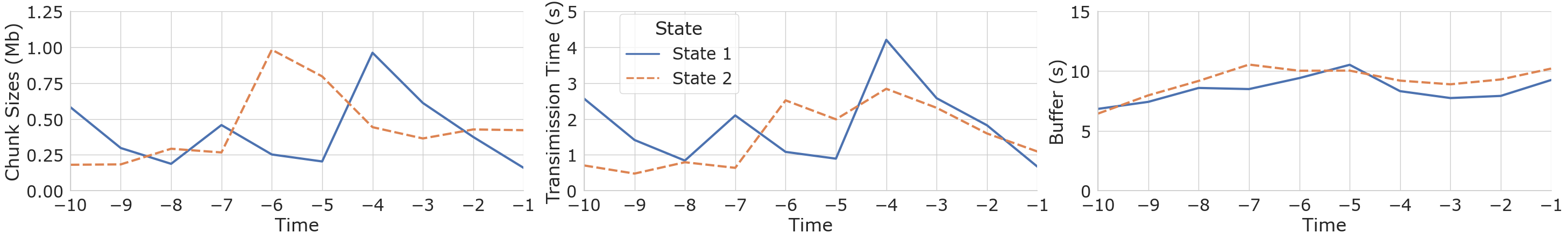}
	\caption{(a) History of States: Visualization of the two motivating states, $S_1$  and $S_2$, with a similar history.}
    \label{fig:motivaton_states}
\end{subfigure}
\begin{subfigure}[t]{0.49\linewidth}
	\centering
	\includegraphics[width=0.87\linewidth]{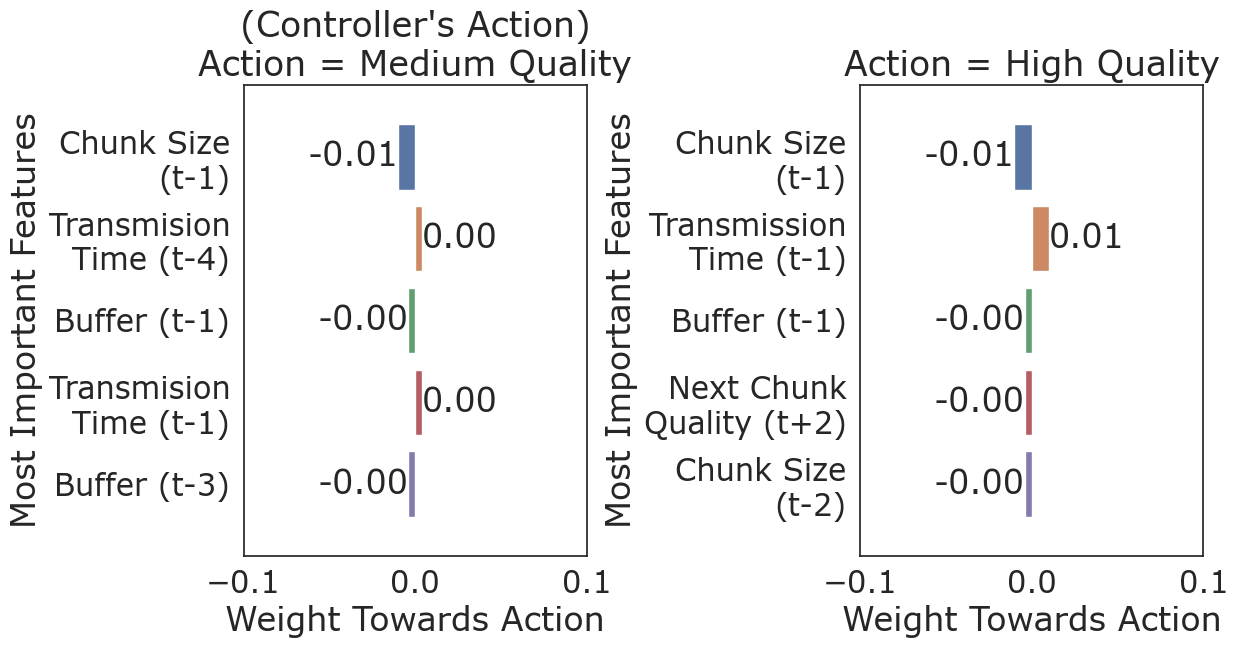}
	\caption{(b) LIME's explanation for $S_1$ showing the key features for the medium and high quality actions.}
 \label{fig:motivation_state1}
\end{subfigure}
\hspace{0.1cm}
\begin{subfigure}[t]{0.49\linewidth}
	\centering
	\includegraphics[width=0.87\linewidth]{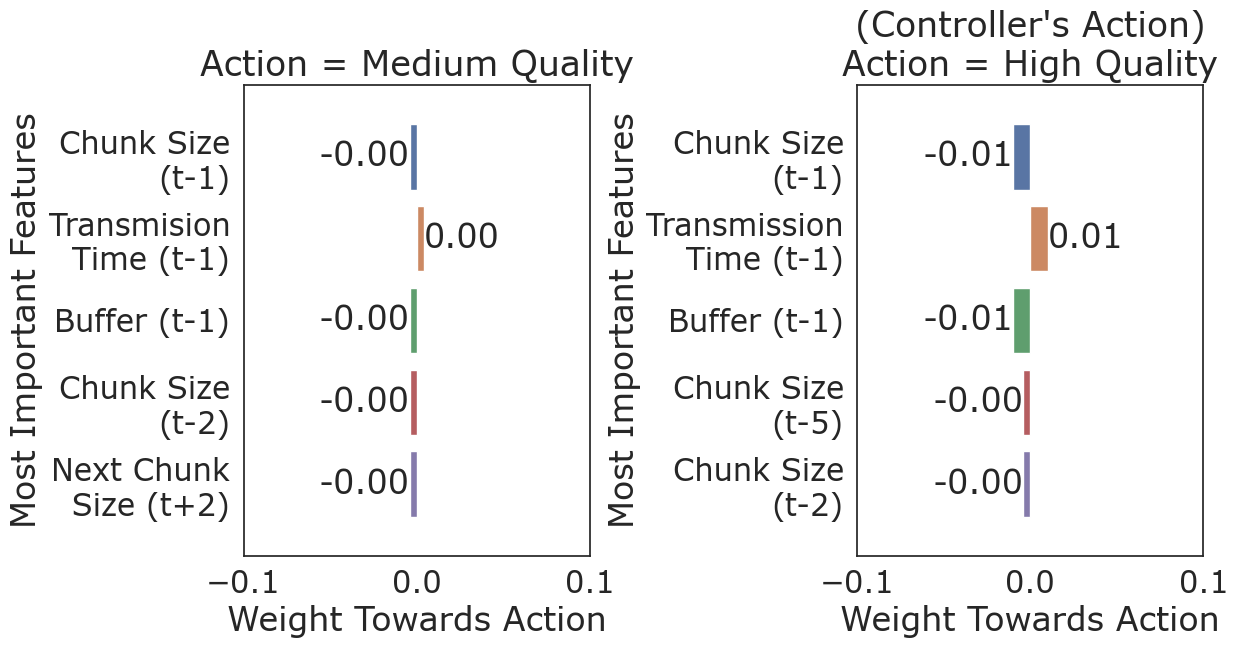}
	\caption{(c) LIME's explanation for $S_2$ showing the key features for the medium and high quality actions. }
    \label{fig:motivation_state2}
\end{subfigure}

\caption{LIME\cite{ribeiro2016should}'s explanation for the motivating states. We observe that in both actions and in both states, LIME presents a similar explanation: recent transmission times, chunk sizes, and buffer are top features.}
\label{fig:motivation}
\end{figure*}

\section{Introduction}
Deep Reinforcement Learning (DRL) outperforms manual heuristics in a broad range of input-driven tasks, including congestion control~\cite{aurora}, adaptive bitrate streaming~\cite{pensieve}, and network traffic optimization~\cite{auto}, among others. However, due to the difficulty in interpreting, debugging, and trusting~\cite{metis}, network operators are often hesitant to deploy them outside lab settings. The domain of explainable reinforcement learning (XRL) seeks to bridge this gap.
 
The field of Explainable Reinforcement Learning, which facilitates human understanding of a model's decision-making process~\cite{burkart2021survey}, can be categorized into two main areas: feature-based and future-based. The feature-based approach applies established techniques from Explainable AI (XAI) in supervised learning to DRL, adapting commonly used post-hoc explainers like saliency maps~\citep{zahavy2016graying, iyer2018transparency, greydanus2018visualizing, puri2019explain} and model distillation~\citep{bastani2018verifiable, verma2018programmatically, zhang2020interpretable} techniques. However, these methods can fail to provide a comprehensive view or explain specific behaviors, because they do not provide insight into the controller's view of the future.

More recently, there is a growing interest in \textit{future-based} explainers, which generate explanations based on a controller's forward-looking view. These explainers detail the effects of a controller's decisions using either future rewards~\cite{juozapaitis2019explainable}  or goals~\cite{van2018contrastive, yau2020did, cruz2021explainable}. However, they necessitate either extensive alterations to the agent~\cite{juozapaitis2019explainable} or precise environment modeling~\cite{van2018contrastive, yau2020did, cruz2021explainable}, both of which pose significant challenges in input-driven environments. Controllers in this domain cannot tolerate performance degradation and are deployed in highly variable real-world conditions.

We present \fmname, a model-agnostic, posthoc explainability framework that generates high-fidelity future-based explanations in input-driven environments. \fmname does not require any modifications to the controller; hence, it is easily deployable and does not sacrifice controller performance. \fmname generates succinct explanations by decomposing the controller's reward function into individual components and using them as the basis for explanations. Reward functions in input-driven environments are naturally decomposable, where the reward components represent key performance metrics. Hence, explanations generated by \fmname are meaningful to operators. For example, in DRL-based congestion control~\cite{aurora}, there are three reward components: throughput, loss, and latency.

We formulate the explainability problem as generating decomposed future returns, given a state and an action. We propose a streamlined model-agnostic supervised learning approach to estimate these returns using an on-policy action-value function. Our solution generalizes across both discrete and continuous control environments. We evaluate our framework on multiple real-world input-driven environments, showing its ability to generate \emph{succinct, high-fidelity} explanations \emph{efficiently}. We demonstrate the usefulness of \fmname's explanations by providing insights when feature-based explainers find it challenging, and powering network observability and guided reward design.

In summary, we make the following key contributions: 
\begin{itemize}[leftmargin=*,nolistsep]
    \item We propose a new notion of future-based explanations in input-driven environments.
    \item We exploit the decomposable and dense reward functions in input-driven environments and formulate a supervised learning problem to estimate future-based explanations.
    \item We demonstrate the practical use of our ideas in several systems applications across input-driven environments.
\end{itemize}

\begin{figure*}[t]
\centering
\begin{subfigure}[t]{0.49\linewidth}
	\centering
	\includegraphics[width=0.825\linewidth]{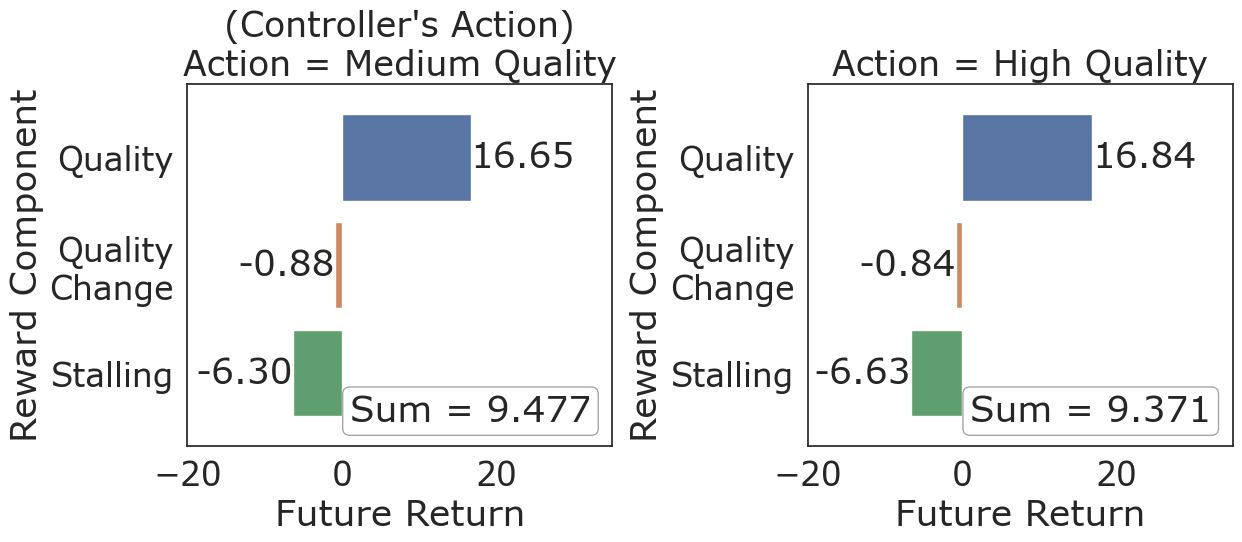}\caption{ (a) CrystalBox's explanation for $S_1$.}
 \label{fig:motivation_explanation_consistent}
\end{subfigure}
\hspace{0.1cm}
\begin{subfigure}[t]{0.49\linewidth}
	\centering
	\includegraphics[width=0.825\linewidth]{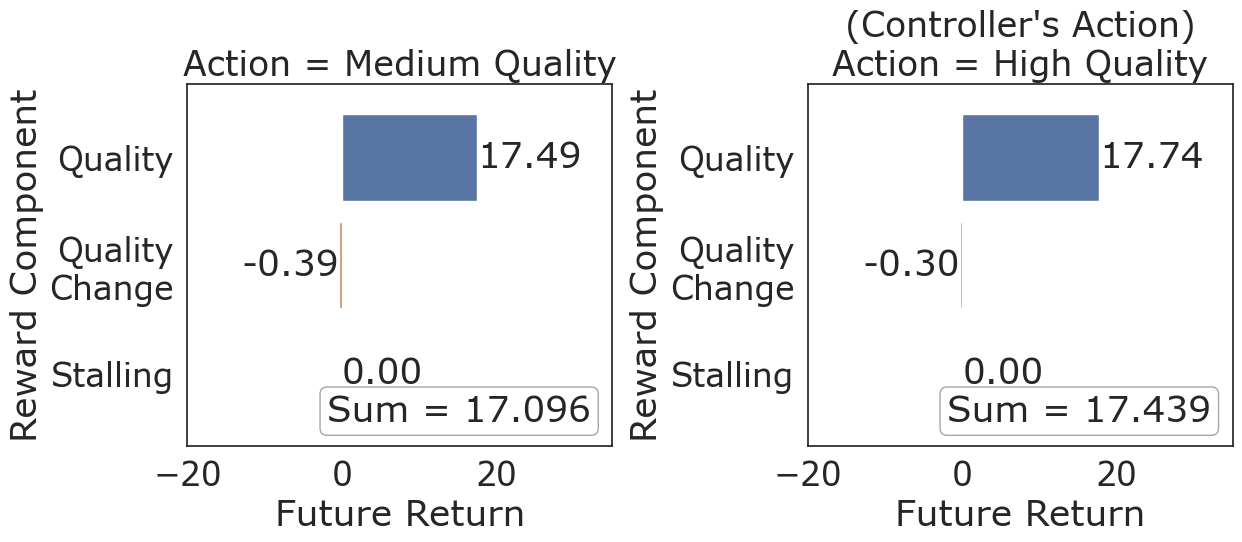}\caption{(b) CrystalBox's explanation for $S_2$.}
 \label{fig:motivation_explanation_recent}
\end{subfigure}

\caption{
CrystalBox's explanation for the two motivating states in Section~\ref{sec:motivation}. CrystalBox allows us to understand why the controller's actions are appropriate in both states by letting us compare their decomposed future returns to those of alternatives.}
\label{fig:crystlbox_motivation_explanation}
\end{figure*}

\section{Motivation}
\label{sec:motivation}

\label{subsec:motivexm}
We use Adaptive Bitrate Streaming (ABR) as a representative input-driven application to understand explainability. In ABR, two entities communicate: a client streaming a video over the Internet, and a server delivering it. The video is usually divided into small, seconds-long chunks, encoded in advance at various bitrates. The ABR controller's goal is to maximize the Quality of Experience (QoE) for the client by selecting the optimal bitrate for subsequent video chunks based on network conditions. QoE is often defined as a linear combination rewarding higher quality and penalizing both quality changes and stalling~\citep{mok2011measuring}.

Let us consider two states, $S_1$ and $S_2$. We visualize them in Figure~\ref{fig:motivaton_states} showing past data like chunk sizes, transmission time, and buffer. The state histories of $S_1$ and $S_2$ are quite similar, with fluctuations in chunk sizes and transmission time, while the client's buffer remains constant. Yet, the DRL controller picks medium-quality bitrate for $S_1$ and high quality bitrate for $S_2$. In this scenario, the network operator seeks to gain two insights to answer standard explainability questions~\cite{doshi2017accountability, van2018contrastive, mittelstadt2019explaining, miller2019explanation}.

\parab{$Q1$: Contrastive explanation within a state}. First, the operator seeks understanding within a single state. This gives rise to question $Q1$: `What causes the controller to choose one action over another within the same state?'\label{qone}

\parab{$Q2$: Contrastive explanation across states}. Second, the operator aims to understand the differing actions of two strikingly similar states. They ask: `Why does the controller choose different actions in similar states?'\label{qtwo}

\noindent We start by analyzing existing explainability frameworks.

\noindent\textbf{Feature-based approach}. 
We choose LIME, a widely employed explainer~\cite{ribeiro2016should}, as a representative feature-based explainer and discuss its ability to answer $Q1$ and $Q2$. LIME is a model-agnostic, local explainer that creates a linear explanation pinpointing the top features driving a given action from an input state. Figure~\ref{fig:motivation_state1} shows the explanations generated by LIME in the state $S_1$ for the two actions medium and high quality bitrates, and \ref{fig:motivation_state2} shows the same for the state $S_2$. 

\textit{Answering $Q1$}. Consider Figure~\ref{fig:motivation_state1} that shows results for $S_1$ and two actions: medium (left plot) and high (right plot) quality bitrates. LIME identifies the top features as the last few values of the chunk sizes, transmission times, and buffer. However, these features are the same for both actions. This leaves no way for the operator to gain an understanding of why the controller picks the top action in this state. The same observation also holds for $S_2$ (Figure~\ref{fig:motivation_state2}).

\textit{Answering $Q2$.} We recall that in these states, the controller opts for medium quality and high quality bitrates, respectively. Looking at LIME's top features in states $S_1$ (Figure~\ref{fig:motivation_state1}) and $S_2$ (Figure~\ref{fig:motivation_state2}) for the preferred actions, the same features appear in both explanations. Despite different preferred actions, LIME identifies similar key features for these decisions. This leaves us questioning why the controller opts for medium quality in one state and high quality in another. LIME falls short in explaining this discrepancy for $Q2$. We hypothesize that the same result holds for other feature-based explainers as they only use the state features.

For a comprehensive answer to these questions, it is necessary to look beyond the features and design explanations with details about the consequences of each action, as the controller inherently selects actions to maximize \textit{future} returns. So, fully understanding the controller's decision-making requires a forward-looking perspective.

\section{Input-Driven Environments}
Input-driven environments~\cite{mao2018variance} represent a rich class of environments reflecting dynamics found in computer systems. Their dynamics are fundamentally different from traditional RL environments. We highlight two of their characteristics: their dependence on input traces and their naturally decomposable reward functions.
\label{subsec:environments}

In input-driven environments, the system conditions are non-deterministic and constitute the main source of uncertainty. They determine how the environment behaves in response to the controller's actions. For instance, in network traffic engineering, network demand influences whether congestion occurs on specific paths. These conditions are referred to as ``inputs''~\cite{mao2018variance}, and the environments with inputs are called input-driven environments. Modeling these inputs remains an open research problem~\cite{park}. In contrast, the presence of a stand-alone simulation environment is often assumed in traditional DRL problems~\cite{gym,beattie2016deepmind}.

A common theme across many of the DRL-based controllers for input-driven environments is a naturally decomposable reward function~\cite{park,aurora,krishnan2018learning}. This is because control in these environments involves optimization across multiple key objectives, which are often linearly combined, with each reward component representing a different objective.

\parab{Input-Driven DRL Environments}.
DRL agents usually train in simulators that mimic real systems. These simulators take the state $s_t$ and action $a_t$ and generate the subsequent state $s_{t+1}$ and reward $r_t$. But, in input-dependent environments, $s_{t+1}$ and $r_t$ depend also on the input value. Therefore, the simulator must also capture the inputs.

Modeling the process behind these inputs can be challenging, as it can involve emulating environments such as the wide-area Internet. To circumvent this issue, state-of-the-art DRL solutions for input-driven environments do not model the complex input process. Instead, during training, they replay traces (or logged runs) gathered from real systems~\cite{park}. In the beginning, the simulator selects a specific trace from the given dataset and generates the next state $s_{t+1}$ by simply looking up the subsequent logged trace value. Note that these traces are not accessible to the agent in advance and are only available during training.

\parab{Formalization}. 
We examine an Input-Driven Markov Decision Process, defined in \cite{mao2018variance}. An Input-Driven MDP is expressed by the tuple $(S, A, Z, P_s, P_z, r, \gamma)$, which represents the set of states ($S$), set of actions ($A$), set of training input-traces ($Z$), state transition function ($P_s$), input transition function ($P_z$), reward function ($r$), and discount ($\gamma$) respectively.

The state transition function $P_s(s_{t+1} | s_t, a_t, z_{t+1})$ defines the probability distribution over the next state given the current state, action, and future input-value ($z_{t+1}$). The function $P_z(z_{t+1} | z_{t})$ defines the likelihood of the next input value given the current one. The combined transition function is $P_s(s_{t+1} | s_t, a_t, z_{t+1}) P_z(z_{t+1} | z_{t})$.

Given $z_t$, obtaining $z_{t+1}$ through $P_z$ 
is an open problem~\cite{park}, as mentioned above. 
During training, this problem is circumvented by selecting a specific $z = [..,z_{t}, z_{t+1},..]$ from the set of traces $Z$ and accessing the subsequent value $z_{t+1}$ until the end of the trace. 
While this method works for training, it is not applicable for computing $P_z$  \emph{outside} of training in algorithms such as Monte Carlo tree search~\cite{browne2012survey} and model predictive control~\cite{holkar2010overview}. 
This is because information about future network conditions, such as throughput values in our motivating example, is not available then.

\section{Future-Based Explanations}

\begin{figure*}[!t]
\centering
\begin{subfigure}[c]{0.445\textwidth}
    \centering
    \includegraphics[width=\textwidth]{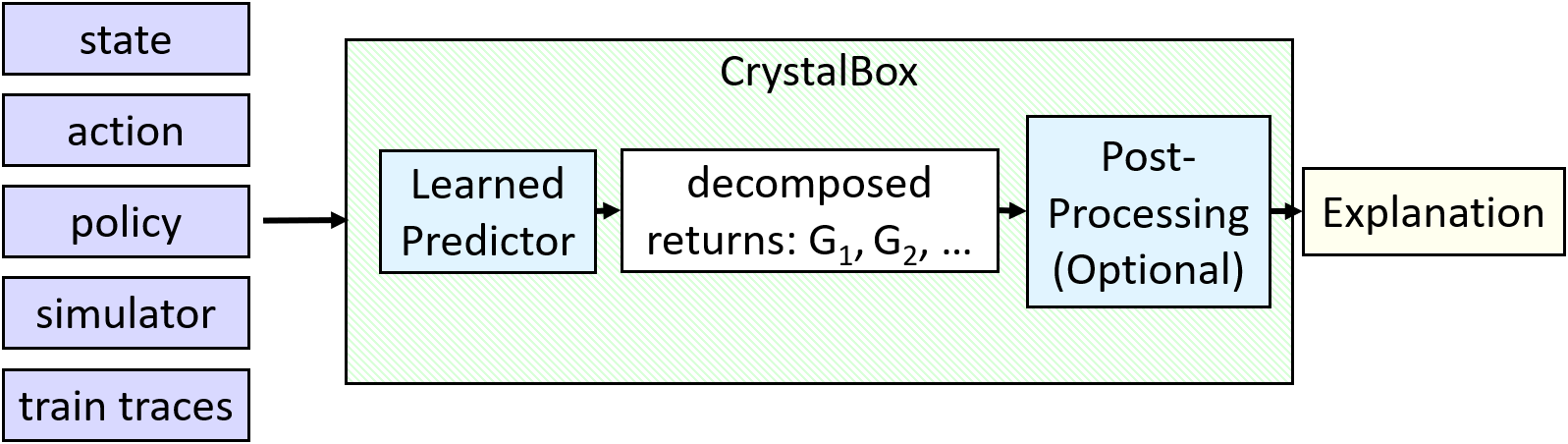}
    \caption{(a)  Overview of CrystalBox.}
    \label{fig:crystalbox_overview}
\end{subfigure}
\hspace{0.15cm}
\begin{subfigure}[c]{.48\textwidth}
    \centering
    \includegraphics[width=\textwidth]{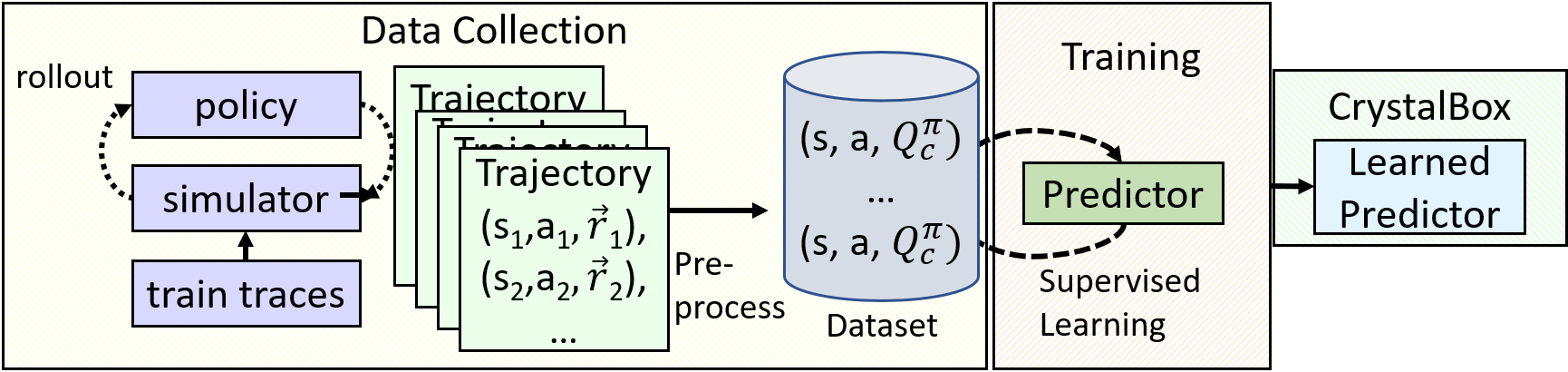}
    \caption{(b) Traning of CrystalBox.}
    \vspace{1mm}
    \label{fig:learned_overview}
\end{subfigure}

\caption{System Diagram of CrystalBox: CrystalBox consists of two components: a learned decomposed returns predictor and a post-processing module. We train a function approximator once to predict the decomposed returns by (i) collecting rollouts of the policy, pre-processing the rollouts to form a dataset, and (ii) employing supervised learning. Once trained, we give the query state and action to this approximator, obtain its predicted decomposed returns, and optionally post-process them.}
\label{fig:crystalbox_diagram}
\end{figure*}

\label{subsec:future_returns}

Future-based explainers elucidate the controller's decision-making process by giving a view into what the controller is optimizing: future performance. Given a state $s_t$ and an action $a_t$, a future-based explainer must provide a forward-looking view into the controller by capturing the consequences of that action in a concise yet expressive language.

We define the problem of generating future-based explanation as computing the decomposed future returns starting from a given state $s_t$ under action $a_t$ and policy $\pi$.
Formally, decomposed future returns are the individual components within the $Q^{\pi}(s_t, a_t) = \sum_{c \in C} Q_c^{\pi}(s_t, a_t)$, where $C$ is the set of reward components in the environment. Here, $Q^\pi$ calculates the expected return for action $a_t$ in state $s_t$ and following policy $\pi$ thereafter. 
For the motivation example, the components in $C$ are quality, quality change, and stalling.

Now, we can define an explanation for a given state, action, and fixed policy as a tuple of return components:
\begin{align}
    {\mathcal X}(\pi, s_t, a_t) = [Q^{\pi}_{c_1}, \ldots, Q^{\pi}_{c_k}],\quad & c_1,\ldots, c_k \in C \label{eq:exp}
\end{align}
Our explanation definition lets us clearly represent future action impact using return components that are typically the key performance metrics. It provides a future-centric view to analyze controller behavior over states and actions.

Let us return to the motivating example (\S~\ref{sec:motivation}). The operator seeks to explain two seemingly similar states, $S_1$ and $S_2$, where the controller chooses different actions (Fig.~\ref{fig:motivation}) and feature-based methods were not able to provide meaningful explanations. We answer these questions with future-based explanations generated by our framework, \fmname (\S~\ref{subsec:crystalbox}).

\textit{Answering $Q1$}. 
Consider Fig.~\ref{fig:motivation_explanation_consistent} that provides an explanation for $S_1$ under both medium and high quality bitrates. The bars represent the positive or negative contributions of each performance metric to the total return. As we can see from the disaggregated performance metrics,
`medium quality' action is preferred as it leads to less stalling and has a higher overall return. Similarly, in
$S_2$ (Fig.~\ref{fig:motivation_explanation_recent}), the disaggregated metrics suggest that the `high quality' action was chosen due to better quality (blue bar) of the streamed video.

\textit{Answering $Q2$}. The operator examines explanations for controller actions in both $S_1$ and $S_2$. They recognize that the expected stalling penalty in $S_1$ steers the preference towards medium quality, while the absence of such penalty in $S_2$ leads to the choice of high quality.
Our example demonstrates that decomposed future returns serve as a useful technique for comparing both actions and states. They enable the operator to gain a comprehensive understanding of the agent's decision-making process.

The next question is how to compute these decomposed future returns. To answer this question let us first formally define return components $Q^{\pi}_{c}(s_t, a_t)$, $\forall c \in C$:
\begin{equation}    
\label{decomposed_q}
\begin{aligned}
    &Q^{\pi}_{c}(s_t, a_t) = r_{c}(s_t, a_t) +  \gamma Q^{\pi}_{c}(s_{t+1}, \pi(s_{t+1})) \Leftrightarrow \\
    &Q^{\pi}_{c}(s_t, a_t) =  r_{c}(s_t, a_t)  + \\
    &\E_{s_{t+1}, a_{t+1}, ...\sim \pi, P_s, P_z} \sum_{\Delta t = 1}^{\infty} [\gamma^{\Delta t} r_{c}(s_{t + \Delta t}, a_{t + \Delta t})],
\end{aligned}
\end{equation}
where $r_{c}(s_t, a_t)$ is the reward value of component $c$ earned by the controller for taking action $a_t$ in state $s_t$. Therefore, to generate explanations, we need to compute this equation.  

Unfortunately, decomposed future returns cannot be directly calculated because, as we see in Eq.~\ref{decomposed_q}, calculating the $Q^\pi_c$ function involves computing $P_z$ for the expected next state. However, $P_z$ cannot be computed outside of training, as explained in Section~\ref{subsec:environments}. There can of course be an approximation for $P_z$. But, obtaining low-bias and low-variance approximation is not currently possible~\cite{park}.

To overcome this limitation, we propose to directly approximate these $Q^\pi_c$ components from examples, using training input traces $Z$. This approach enables us to bypass the need for $P_z$, making it feasible to generate explanations outside of training. In contrast to sparse-reward environments such as mazes, chess, and go, where such approximations may face considerable variance challenges~\cite{sutton2018reinforcement}, input-driven environments provide informative, consistent, non-zero rewards at each step. This feature makes our approximation practicable in input-driven contexts.
Here, such approximation can be done either through a sampling-based or a learning-based procedure. We designed and evaluated several sampling-based approaches (\S~\ref{subsec:sampling}) and found their performance unsatisfactory (\S~\ref{sec:evaluation}).

\section{CrystalBox}

\label{subsec:crystalbox}

We define a new supervised learning problem to learn decomposed future returns  $Q^{\pi}_c$ in input-driven environments.
We exploit the key insight that  $Q^{\pi}_c(s_t, a_t)$ form a function that can be directly parameterized and learned. Formulating a supervised learning problem requires us to design (i) how we obtain the samples, (ii) the function approximator architecture, and (iii) the loss function. Fig.~\ref{fig:crystalbox_overview} shows the high-level flow of our approach and Fig.~\ref{fig:learned_overview} focuses on the details of the learning procedure.

First, we propose a procedure to obtain samples of $Q^\pi_c$. From Eq.~\ref{decomposed_q}, we see that $Q^\pi_c$ depends on the combined transition function of the environment. This combined transition function forms a power distribution involving the variables $P_z$ and $P_s$. $P_z$ cannot be approximated with low bias or variance, leading to high bias estimations of $Q^\pi_c$. To solve this problem, we circumvent approximating $P_z$ by instead using its unbiased samples. These unbiased samples are exactly the traces from the training set of traces $Z$.
We replay these traces with a given policy $\pi$ to obtain estimates of $Q^\pi_c$ (See Fig.~\ref{fig:learned_overview} `Data Collection' block) using Monte-Carlo rollouts~\cite{sutton2018reinforcement}. 
Second, we define our function approximator architecture with parameters $\theta$. We use a simple split fully connected layers architecture (Appendix~A.5\footnote{This is the extended version of the paper~\cite{patel2024crystalbox}}) to jointly learn the return components. Finally, we define a loss for each individual component $c$ and take a weighted sum as the overall loss. We use standard Mean Squared Error for our individual components.   

While our formulation is unbiased, it can still have high variance due to its dependence on $P_z$. As a result, $Q^\pi_c$ can have high variance in realistic settings, requiring a large number of samples to achieve sufficient coverage.  We employ two generic strategies to overcome this. First, we notice that reusing the embeddings $\phi(s_t)$ from the policy instead of the states $s_t$ directly helps to reduce the variance. 
The embeddings $\phi(s_t)$ capture important concepts about states, reducing the variance at the early stages of predictor learning. Second, we collect two kinds of rollouts: on-policy and exploratory (see Fig.~\ref{fig:learned_overview}). On-policy rollouts strictly follow the policy, whereas exploratory rollouts start with an explorative action. This approach enables \fmname to achieve higher coverage of $Q^\pi_c$ for all actions. 

Overall, \fmname will converge to the true $Q^\pi$ function, reflecting the policy's performance. This stems from the fact that our formulation is a special instance of the Monte Carlo Policy Evaluation algorithm~\cite{silver2015, sutton2018reinforcement} for estimating $Q^\pi_{\theta}$. Here, $Q^\pi_{\theta}$ is split into smaller parts $Q^\pi_{\theta, c}$, summable to the original. Consequently, the Monte Carlo Policy Evaluation's correctness proof applies. 

We highlight \fmname applies to any fixed policy $\pi$, even if $\pi$ is non-deterministic or has continuous action space. All that is required is access to a simulator with training traces, which is publicly available for most input-driven RL environments~\cite{park}.

\section{Comparing Explanations}

In this section, we give an overview of metrics and baselines that we use for evaluating CrystalBox.

\subsection{Quality of explanations}
\label{subsec:fidelity}
In standard explainability workflow, an explainer takes a complex function $f(x)$ and produces an interpretable approximation $g(x)$. To measure the quality of the approximation, the commonly used fidelity metric $FD = \lVert f(x) - g(x) \rVert, x\in {\mathcal D}$ measures how closely the approximation follows the original function under a region of interest ${\mathcal D}$.

This metric can rather directly be applied to our explanations. As above, we have the complex function $Q^{\pi}_c$, one per each component $c$ (\S~\ref{subsec:future_returns}). \fmname outputs an approximation, $\mathrm{Pred}(Q^{\pi}_c)$, that also serves as an explanation. Hence, the fidelity metric is defined as a norm between a complex function and its approximation: 
\begin{align}
    FD_c = \lVert Q^{\pi}_c - \mathrm{Pred}(Q^{\pi}_c) \rVert, \forall c \in C.
\end{align}
In our experiments, we use the $L_2$ norm. However, there is one distinction to discuss. Unlike standard settings, $Q^{\pi}_c$ is neither explicitly given to us nor can be computed outside of training (\S~\ref{subsec:future_returns}). Thus, the best we can do to simulate real-world evaluation is to use a held-out set of traces $Z'$ and rollout policy $\pi$ to obtain $\estq{Q}_c(s_t, a_t)$ under this held-out set.

\subsection{Sampling Techniques}
\label{subsec:sampling}
As we mentioned in Section~\ref{subsec:future_returns}, an alternative way to compute $Q^\pi_c$ components by using sampling-based techniques that we discuss next. 
These techniques also serve as natural baselines for \fmname.

We estimate the individual components of $Q^\pi(s_t, a_t)$  empirically by averaging over the outcomes of running simulations starting from $s_t$ and taking the action $a_t$.
To approximate $Q_c^{\pi}(s_t, a_t)$, we must sample potential futures for the state $s_t$. 
However, in input-driven environments, neither $z_{t+1}$ nor $P_z$ are available to sample the state outside training (\S~\ref{subsec:environments}). Hence, we sample potential $z$ futures from our training trace dataset to best guess 
these potential futures. We consider two possible variants here.

\parab{Naive Sampling} A simple strategy for sampling involves uniformly random sampling $z$ from our training set $Z$. We observed that this method suffers from low accuracy (see \S~\ref{sec:evaluation}) due to its dependence on the dataset distribution, which can be unbalanced in input-driven applications. Dominant traces often do not cover all relevant scenarios (see Fig. 12 and 13 in Appendix~A.1).  

\parab{Distribution-Aware Sampling}. We explore one avenue to improve the accuracy of naive sampling: making our sampling procedure distribution aware. We approximate $P(z_{t+1} | s_t)$ by clustering all traces in the dataset, observing the input values from $s_t$, and mapping it to the nearest cluster. A trace within that cluster is then randomly sampled. This conditioning improves naive sampling (see \S~\ref{sec:evaluation}).

\begin{figure*}[!th]
\centering
\begin{subfigure}[t]{.99\textwidth}
    \centering
    \includegraphics[width=0.45\textwidth]{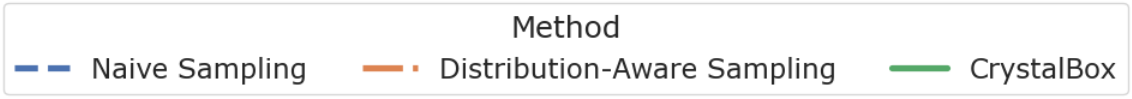}
\end{subfigure}
\newline
\begin{subfigure}[t]{.49\textwidth}
    \centering
    \includegraphics[width=\textwidth]{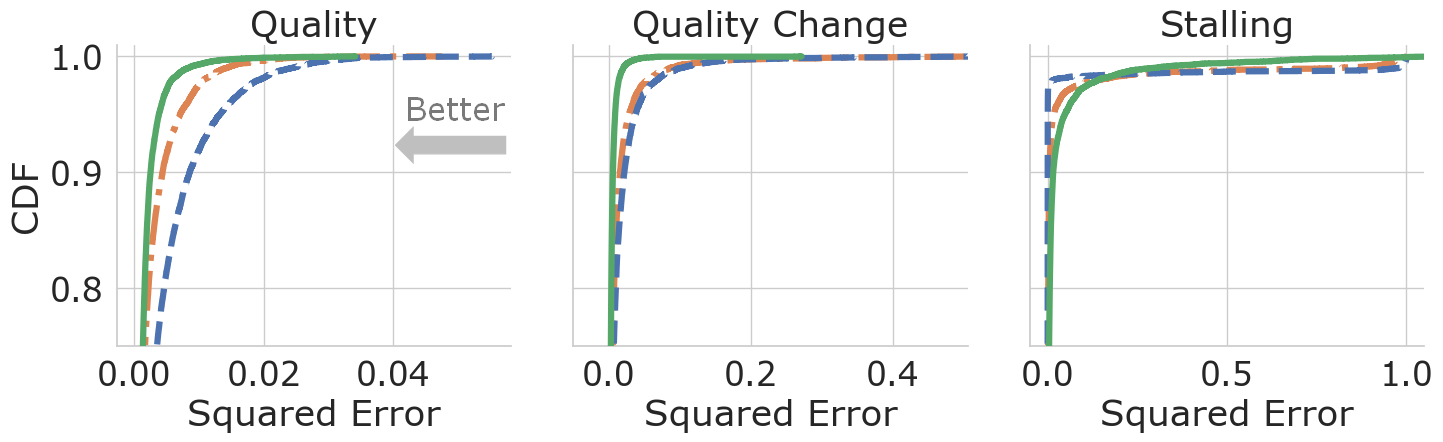}
    \caption{(a) Adaptive Bitrate Streaming: factual actions}
    \label{fig:abr_factual_mse}
\end{subfigure}
\begin{subfigure}[t]{.49\textwidth}
    \centering
    \includegraphics[width=\textwidth]{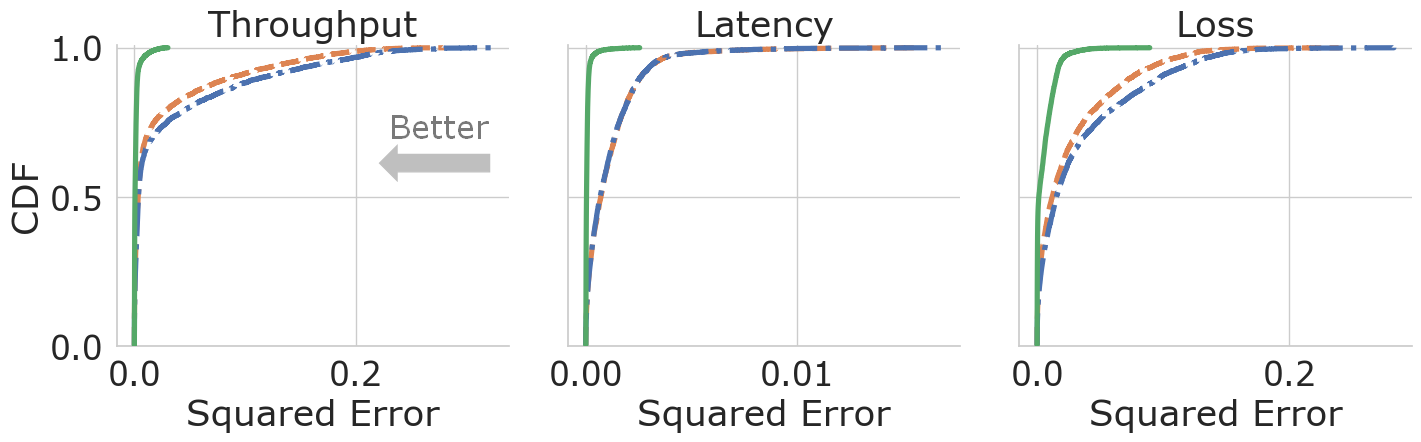}
    \caption{(b) Congestion Control: factual actions}
    \label{fig:cc_factual_mse}
\end{subfigure}
\begin{subfigure}[t]{.49\textwidth}
    \centering
    \includegraphics[width=\textwidth]{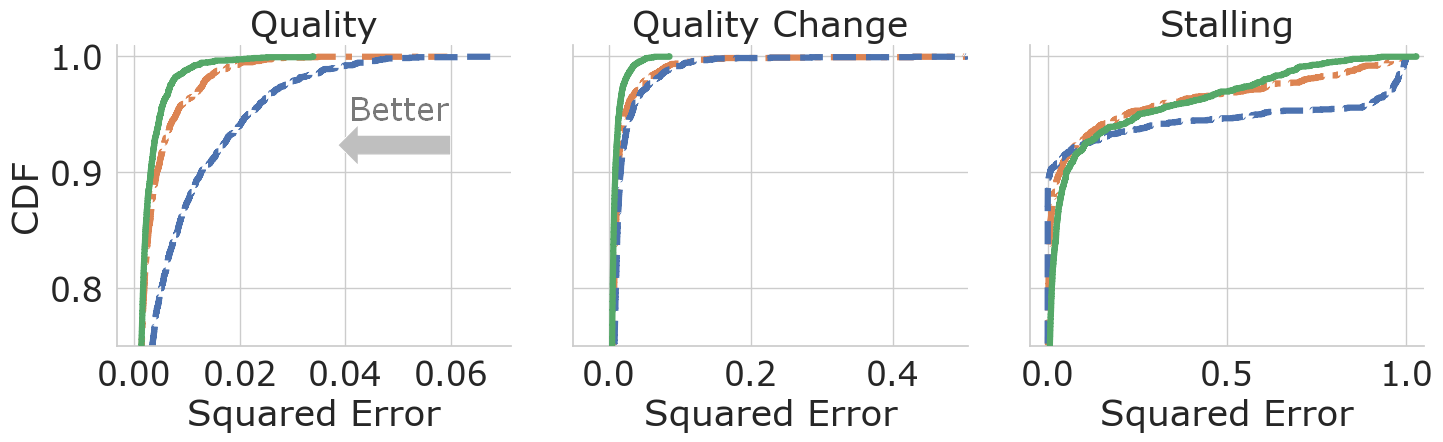}
    \caption{(c) Adaptive Bitrate Streaming: counterfactual actions}
    \label{fig:abr_counter_factual_mse}
\end{subfigure}
\begin{subfigure}[t]{.49\textwidth}
    \centering
    \includegraphics[width=\textwidth]{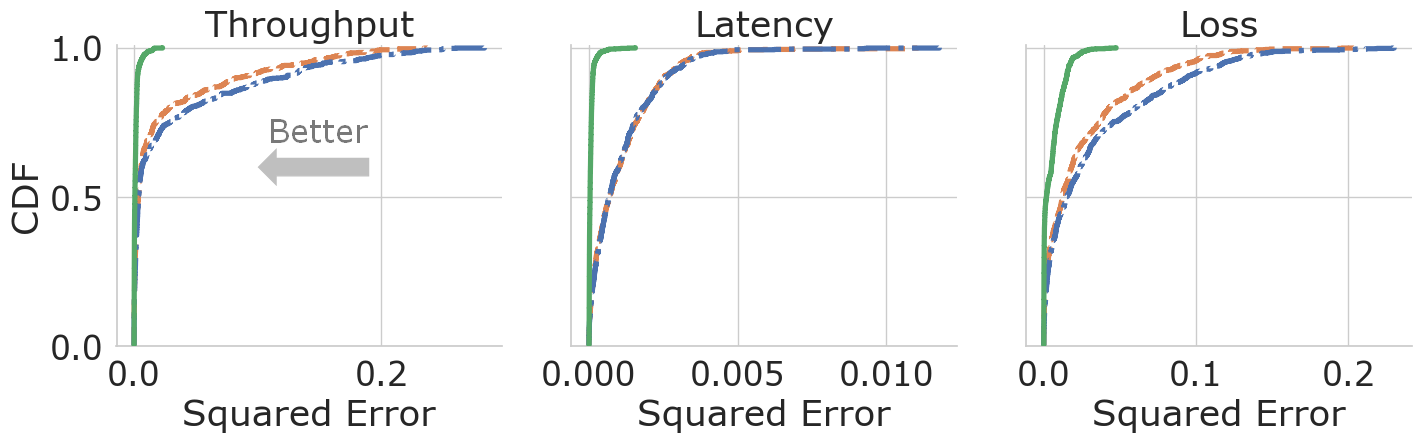}
    \caption{(d) Congestion Control: counterfactual actions}
    \label{fig:cc_counter_factual_mse}
\end{subfigure}

\caption{Fidelity Evaluation of CrystalBox: Distribution of Squared Error of different methods to the ground truth samples in ABR and CC. 
CrystalBox offers predictions with the lowest error to the ground truth in all three return components of both environments, for both factual and counterfactual actions. Note that the values of all the returns are scaled to the range [0, 1] before being measured for error. The y-axis in results for ABR is adjusted due to the inherent tail-ended nature of ABR.} 
\label{fig:main_mse_results}
\end{figure*}

\section{Experiments}
\label{sec:evaluation}
We present an evaluation of \fmname\footnote{\url{https://github.com/sagar-pa/crystalbox}}. We aim to answer the following questions: Does CrystalBox produce high-fidelity explanations? Is CrystalBox efficient? Does joint training help? What applications can CrystalBox enable?

\subsection{Experimental Setup}
We evaluate CrystalBox using two input-driven environments: Adaptive Bitrate Streaming (ABR) and Congestion Control (CC). In ABR, the controller selects the video quality for an online stream presented to a user. The reward for the controller is the user's quality of experience, quantified as a weighted sum of quality, quality change, and stalling. In CC, the controller manages the Internet traffic of a connection between a sender and a receiver by adjusting the sending rate of the sender. The reward here is a weighted sum of throughput, latency, and loss.
ABR has \emph{discrete} actions and CC has \emph{continuous} actions. 

We experiment with the publicly available ABR controller maguro~\cite{patel2023plume} deployed on the Puffer Platform~\cite{fugu}. It is the best ABR controller on Puffer. For Congestion Control, we borrow the DRL controller Aurora~\cite{aurora}. 

In all of our evaluations, we use a held-out set of traces $Z'$. We use it to evaluate fidelity, controller performance, and network observability.

\subsection{Fidelity Evaluation}

We evaluate \fmname under two classes of actions: actions that the policy takes (factual actions), and actions that the policy does not take (counterfactual actions). That enables a wide range of explainability queries, including `Why A?' and `Why A and not B?'. We compare predictions of \fmname with sampling techniques (\S~\ref{subsec:sampling}) for each return component $Q_c^{\pi}$ using the fidelity metric (\S~\ref{subsec:fidelity}).

In Figure~\ref{fig:main_mse_results}, we show the error of returns predicted by CrystalBox and sampling baselines. We see that CrystalBox outperforms both of the sampling approaches in generating high-fidelity predictions for all three return components across both environments, under factual (Fig.~\ref{fig:abr_factual_mse},~\ref{fig:abr_counter_factual_mse}) and counterfactual (Fig.~\ref{fig:cc_factual_mse},~\ref{fig:cc_counter_factual_mse}) actions. Note that despite ABR's discrete action space and CC's continuous one, CrystalBox offers high-fidelity explanations in each case. We believe that this is due to CrystalBox's estimates being unbiased by construction, unlike the ones from sampling baselines.

\subsection{CrystalBox Analysis}
\begin{figure*}[t]
\centering
    \begin{subfigure}[t]{.48\textwidth}
    \centering
    \includegraphics[width=0.7\linewidth]{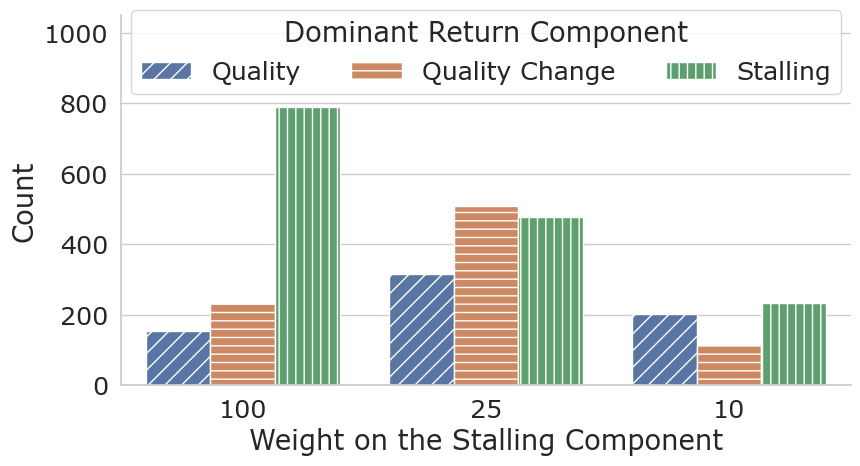}
    \caption{Figure \arabic{figure}: Tuning the weight of Stalling Reward Component in ABR. Here, we employ CrystalBox to explain why the controller chooses to drop bitrate in seemingly good states. We identify the dominant reward components and plot them.}
    \labeltext{\arabic{figure}}{fig:abr_tuning}
    \addtocounter{figure}{1}
    \addtocounter{subfigure}{-1}
    \end{subfigure}
\hspace{0.7cm}
    \begin{subfigure}[t]{.47\textwidth}
    \centering
    \includegraphics[width=0.7\linewidth]{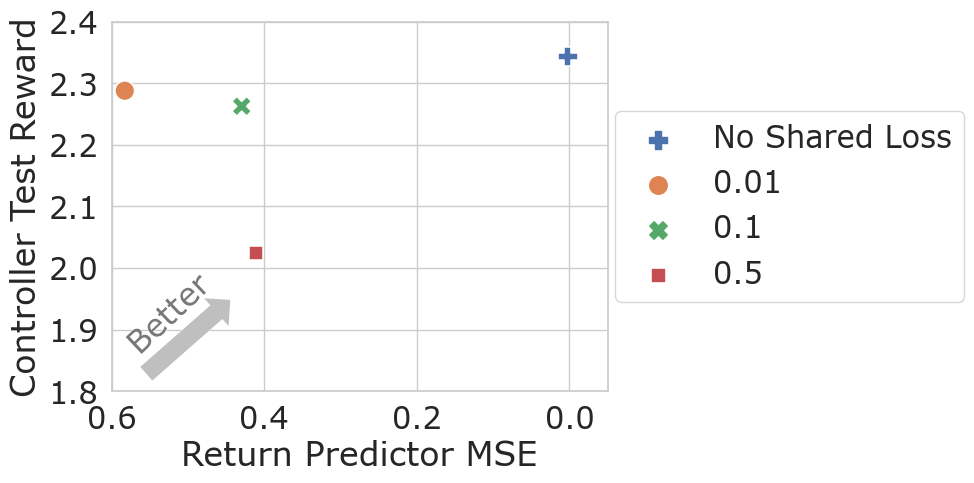}
    \caption{Figure \arabic{figure}: Combining CrystalBox with the agent: We jointly train CrystalBox and the controller. We plot the weight we assign to CrystalBox's loss, from 0.5 of the controller's loss to 0.01.}
    \labeltext{\arabic{figure}}{fig:crystalbox_shared_loss}
    \end{subfigure}
\end{figure*}

\begin{figure*}[!t]
\centering
\begin{subfigure}[t]{.99\textwidth}
    \centering
    \includegraphics[width=0.45\textwidth]{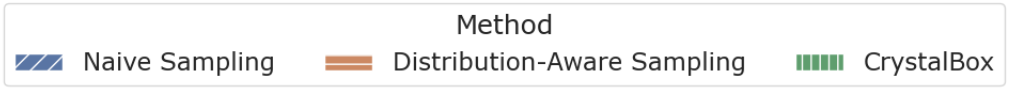}
\end{subfigure}
\newline
\begin{subfigure}[t]{.246\textwidth}
    \centering
    \includegraphics[width=\textwidth]{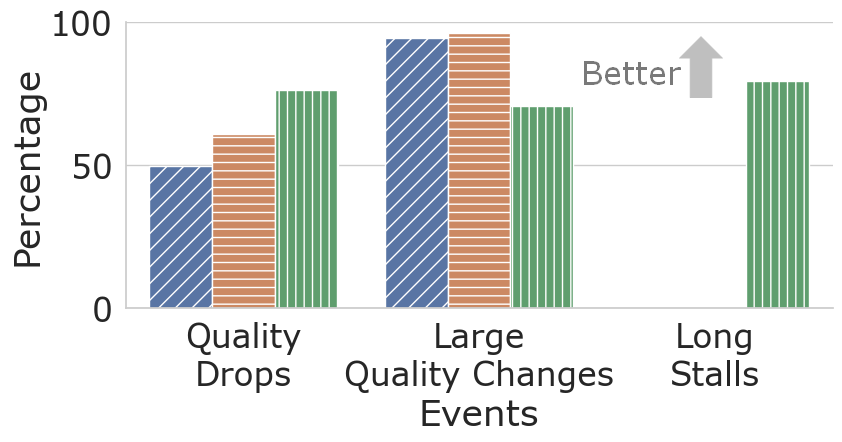}
    \caption{(a) ABR: Recall}
    \label{fig:event_detect_abr_recall}
\end{subfigure}
\begin{subfigure}[t]{.246\textwidth}
    \centering
    \includegraphics[width=\textwidth]{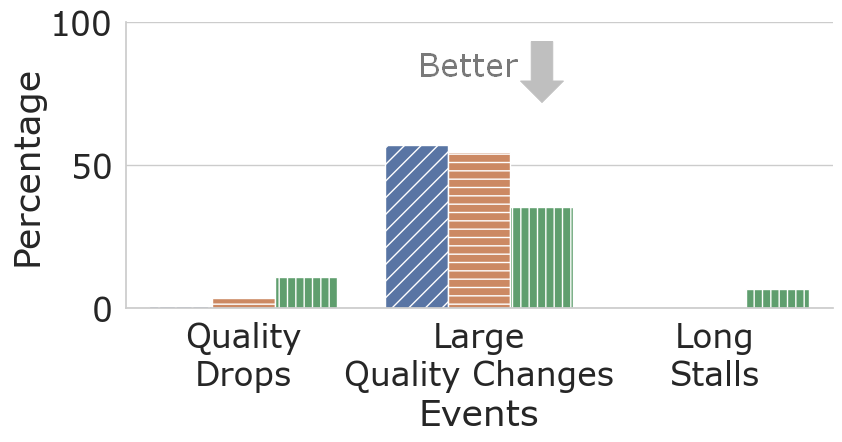}
    \caption{(b) ABR: False Positive Rate}
    \label{fig:event_detect_abr_fp}
\end{subfigure}
\begin{subfigure}[t]{.246\textwidth}
    \centering
    \includegraphics[width=\textwidth]{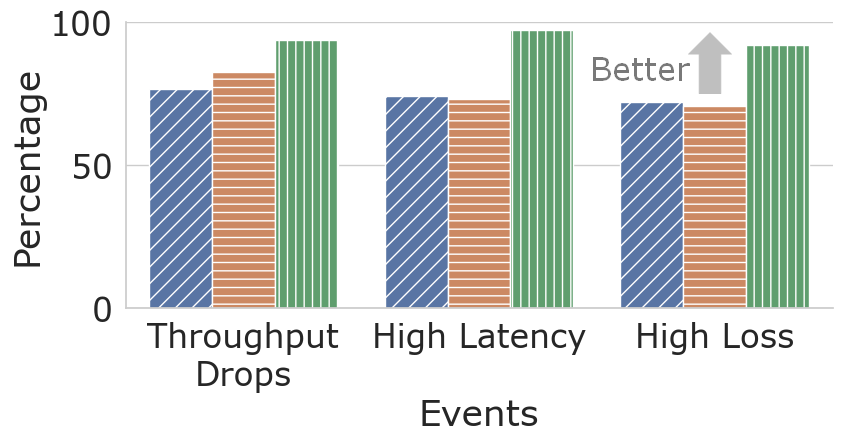}
    \caption{(c) CC: Recall}
    \label{fig:event_detect_cc_recall}
\end{subfigure}
\begin{subfigure}[t]{.246\textwidth}
    \centering
    \includegraphics[width=\textwidth]{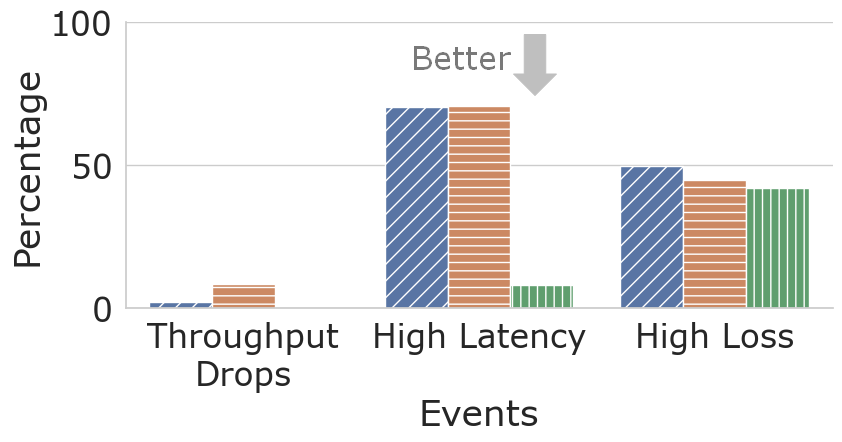}
    \caption{(d) CC: False Positive Rate}
    \label{fig:event_detect_cc_fp}
\end{subfigure}
\caption{Large Performance Drop Event Detection: We analyze the efficacy of different predictors for detecting large performance drops. We identify events happening by detecting if samples of the ground-truth return exceed a threshold.}
\label{fig:event_detection_factual}
\end{figure*}

We turn to present a closer analysis of CrystalBox, focusing on its inference latency and design choices.

\parab{Runtime Analysis}. We analyze the efficiency of CrystalBox by looking at its output latency and contrasting it with sampling-based techniques. We find that in both ABR and CC, CrystalBox has a latency of less than $10$ms, while sampling-based methods have a latency from $50$ms to $250$ms.
Note that this low latency of \fmname enables real-time applications such as networking observability.

\parab{Joint Training Through an Auxiliary Loss Function}. In designing CrystalBox, our priority was to avoid modifying the controller or its training process. This flexibility allows CrystalBox to work across various policies and environments without necessitating redesigns.

An immediate question that may arise is if modifying the agent could yield a better explainer. To explore this, we run an experiment where the controller and explainer are jointly trained without altering the controller's training parameters. We optimize the RL algorithm’s loss and CrystalBox’s loss with a weighted sum. One may postulate this might enhance both the policy and the explainer~\cite{lyle2021effect}.

In Figure~\ref{fig:crystalbox_shared_loss}, we study the shared-training strategy of using a weighted joint loss function in congestion control. Increasing the weight for policy loss improves the policy's quality to nearly match one trained without a predictor, but the quality of \fmname falls as anticipated (see orange dot). Conversely, if we increase the weight for \fmname, the policy quality drops and the predictor does not improve (see red dot). While this may be surprising, we hypothesize that such behavior should be expected.
Recall that samples of $Q^\pi_c$ are unbiased, but have high variance. Therefore, it requires a large number of samples to effectively learn. When we attempt to learn both a policy and a predictor simultaneously, we introduce an additional source of variance, with $Q^\pi_c$ continuously changing with policy changes. As a result, devising a solution for joint training is highly non-trivial. 

\subsection{CrystalBox Applications}
Lastly, we evaluate CrystalBox on concrete applications.

\parab{Network Observability}.
\label{subsec:event_detect_eval}
\fmname can assist an operator in observing system behavior by triggering potential performance degradation alerts. These alerts can help maintain online safety assurance~\cite{rotman2020online} and improve the controller.

Managing data large streams in observability is challenging, leading us to propose a post-processing mechanism. It utilizes a `threshold' concept to demarcate boundaries for binary events. When a threshold is reached, an alert is triggered. Threshold values can be set based on risk and cost. 

We assess the performance of our alert mechanism, treating alerts as event predictors in a binary classification problem. We rollout policy $\pi$ under the held-out trace set $Z'$, considering two kinds of actions: the policy's actions (factual) and alternative actions (counterfactual). Then, we observe if the threshold was reached for the state and action.

Figure~\ref{fig:event_detection_factual} shows our results for ABR and CC under factual actions. Observing the ABR results, we note the following. CrystalBox exhibits high recall and low false-positive rates for both quality drops and long stalls in factual explanations. Sampling-based methods miss all long stalls but detect large quality changes better, albeit with higher false positives. These observations hold for the CC environment. Similar findings for counterfactual actions are in Appendix~A.2.

\parab{Guiding Reward Design}. 
\label{subsec:reward-design}
Fine-tuning the weights of the reward function is a pain point for operators. CrystalBox can help in simplifying this process by letting us narrowly analyze their impact on specific scenarios.

Consider a scenario where the operator is troubleshooting the weights on the reward function of an ABR controller that reduces the bitrate even in seemingly `good' states. To gain insight, we query CrystalBox for these states with two actions: the controller's (where the bitrate drops) and a steady one continuing at the last bitrate. Then, we identify the dominant reward component that pushes the controller to deviate from steady action through the largest absolute difference between the two predicted decomposed future returns.
 
In Figure~\ref{fig:abr_tuning}, we plot the frequency at which each reward component was found to be dominant. The operator first chooses a stall weight of 100 (leftmost bars) and observes that the stalling penalty dominates the decision-making process of the controller. This finding hints to the designer that the weight of the stalling penalty is too high and that they should try 25 (middle bars) or 10 (right bars) where the number of bitrate drops in such states is smaller, and less often motivated by the stalling reward component.

\section{Discussion and Future Work}

Prior work~\cite{juozapaitis2019explainable,anderson2019explaining} introduced a method to generate explanations based on a decomposed deep Q-learning technique for game environments and evaluated this idea on grid-world games. This approach is designed for Q-learning policies, and Q-learning cannot be applied to continuous control input-driven environments. Moreover, they assume access to and the ability to modify the training procedure. In contrast, our approach works for both discrete and continuous control systems without altering the training procedure.

Investigating the use of feature-based techniques for creating interpretable future return predictors represents a compelling research direction. Another potential avenue is to explore whether we can employ future return predictors during policy learning for human-in-the-loop frameworks.

\bibliography{aaai24}

\begin{thebibliography}{39}
\providecommand{\natexlab}[1]{#1}

\bibitem[{Anderson et~al.(2019)Anderson, Dodge, Sadarangani, Juozapaitis, Newman, Irvine, Chattopadhyay, Fern, and Burnett}]{anderson2019explaining}
Anderson, A.; Dodge, J.; Sadarangani, A.; Juozapaitis, Z.; Newman, E.; Irvine, J.; Chattopadhyay, S.; Fern, A.; and Burnett, M. 2019.
\newblock Explaining reinforcement learning to mere mortals: An empirical study.
\newblock \emph{arXiv preprint arXiv:1903.09708}.

\bibitem[{Bastani, Pu, and Solar-Lezama(2018)}]{bastani2018verifiable}
Bastani, O.; Pu, Y.; and Solar-Lezama, A. 2018.
\newblock Verifiable reinforcement learning via policy extraction.
\newblock \emph{Advances in neural information processing systems}, 31.

\bibitem[{Beattie et~al.(2016)Beattie, Leibo, Teplyashin, Ward, Wainwright, K{\"u}ttler, Lefrancq, Green, Vald{\'e}s, Sadik et~al.}]{beattie2016deepmind}
Beattie, C.; Leibo, J.~Z.; Teplyashin, D.; Ward, T.; Wainwright, M.; K{\"u}ttler, H.; Lefrancq, A.; Green, S.; Vald{\'e}s, V.; Sadik, A.; et~al. 2016.
\newblock Deepmind lab.
\newblock \emph{arXiv preprint arXiv:1612.03801}.

\bibitem[{Brockman et~al.(2016)Brockman, Cheung, Pettersson, Schneider, Schulman, Tang, and Zaremba}]{gym}
Brockman, G.; Cheung, V.; Pettersson, L.; Schneider, J.; Schulman, J.; Tang, J.; and Zaremba, W. 2016.
\newblock Openai gym.
\newblock \emph{arXiv preprint arXiv:1606.01540}.

\bibitem[{Browne et~al.(2012)Browne, Powley, Whitehouse, Lucas, Cowling, Rohlfshagen, Tavener, Perez, Samothrakis, and Colton}]{browne2012survey}
Browne, C.~B.; Powley, E.; Whitehouse, D.; Lucas, S.~M.; Cowling, P.~I.; Rohlfshagen, P.; Tavener, S.; Perez, D.; Samothrakis, S.; and Colton, S. 2012.
\newblock A survey of monte carlo tree search methods.
\newblock \emph{IEEE Transactions on Computational Intelligence and AI in games}, 4(1): 1--43.

\bibitem[{Burkart and Huber(2021)}]{burkart2021survey}
Burkart, N.; and Huber, M.~F. 2021.
\newblock A survey on the explainability of supervised machine learning.
\newblock \emph{Journal of Artificial Intelligence Research}, 70: 245--317.

\bibitem[{Chen et~al.(2018)Chen, Lingys, Chen, and Liu}]{auto}
Chen, L.; Lingys, J.; Chen, K.; and Liu, F. 2018.
\newblock Auto: Scaling deep reinforcement learning for datacenter-scale automatic traffic optimization.
\newblock In \emph{Proceedings of the 2018 conference of the ACM special interest group on data communication}, 191--205.

\bibitem[{Cruz et~al.(2021)Cruz, Dazeley, Vamplew, and Moreira}]{cruz2021explainable}
Cruz, F.; Dazeley, R.; Vamplew, P.; and Moreira, I. 2021.
\newblock Explainable robotic systems: Understanding goal-driven actions in a reinforcement learning scenario.
\newblock \emph{Neural Computing and Applications}, 1--18.

\bibitem[{Doshi-Velez et~al.(2017)Doshi-Velez, Kortz, Budish, Bavitz, Gershman, O'Brien, Scott, Schieber, Waldo, Weinberger et~al.}]{doshi2017accountability}
Doshi-Velez, F.; Kortz, M.; Budish, R.; Bavitz, C.; Gershman, S.; O'Brien, D.; Scott, K.; Schieber, S.; Waldo, J.; Weinberger, D.; et~al. 2017.
\newblock Accountability of AI under the law: The role of explanation.
\newblock \emph{arXiv preprint arXiv:1711.01134}.

\bibitem[{Greydanus et~al.(2018)Greydanus, Koul, Dodge, and Fern}]{greydanus2018visualizing}
Greydanus, S.; Koul, A.; Dodge, J.; and Fern, A. 2018.
\newblock Visualizing and understanding atari agents.
\newblock In \emph{International conference on machine learning}, 1792--1801. PMLR.

\bibitem[{Holkar and Waghmare(2010)}]{holkar2010overview}
Holkar, K.; and Waghmare, L.~M. 2010.
\newblock An overview of model predictive control.
\newblock \emph{International Journal of control and automation}, 3(4): 47--63.

\bibitem[{Iyer et~al.(2018)Iyer, Li, Li, Lewis, Sundar, and Sycara}]{iyer2018transparency}
Iyer, R.; Li, Y.; Li, H.; Lewis, M.; Sundar, R.; and Sycara, K. 2018.
\newblock Transparency and explanation in deep reinforcement learning neural networks.
\newblock In \emph{Proceedings of the 2018 AAAI/ACM Conference on AI, Ethics, and Society}, 144--150.

\bibitem[{Jay et~al.(2019)Jay, Rotman, Godfrey, Schapira, and Tamar}]{aurora}
Jay, N.; Rotman, N.; Godfrey, B.; Schapira, M.; and Tamar, A. 2019.
\newblock A deep reinforcement learning perspective on internet congestion control.
\newblock In \emph{International conference on machine learning}, 3050--3059. PMLR.

\bibitem[{Juozapaitis et~al.(2019)Juozapaitis, Koul, Fern, Erwig, and Doshi-Velez}]{juozapaitis2019explainable}
Juozapaitis, Z.; Koul, A.; Fern, A.; Erwig, M.; and Doshi-Velez, F. 2019.
\newblock Explainable reinforcement learning via reward decomposition.
\newblock In \emph{IJCAI/ECAI Workshop on explainable artificial intelligence}.

\bibitem[{Krishnan et~al.(2018)Krishnan, Yang, Goldberg, Hellerstein, and Stoica}]{krishnan2018learning}
Krishnan, S.; Yang, Z.; Goldberg, K.; Hellerstein, J.; and Stoica, I. 2018.
\newblock Learning to optimize join queries with deep reinforcement learning.
\newblock \emph{arXiv preprint arXiv:1808.03196}.

\bibitem[{Langley et~al.(2017)Langley, Riddoch, Wilk, Vicente, Krasic, Zhang, Yang, Kouranov, Swett, Iyengar et~al.}]{langley2017quic}
Langley, A.; Riddoch, A.; Wilk, A.; Vicente, A.; Krasic, C.; Zhang, D.; Yang, F.; Kouranov, F.; Swett, I.; Iyengar, J.; et~al. 2017.
\newblock The quic transport protocol: Design and internet-scale deployment.
\newblock In \emph{Proceedings of the conference of the ACM special interest group on data communication}, 183--196.

\bibitem[{Lyle et~al.(2021)Lyle, Rowland, Ostrovski, and Dabney}]{lyle2021effect}
Lyle, C.; Rowland, M.; Ostrovski, G.; and Dabney, W. 2021.
\newblock On the effect of auxiliary tasks on representation dynamics.
\newblock In \emph{International Conference on Artificial Intelligence and Statistics}, 1--9. PMLR.

\bibitem[{Mao et~al.(2019)Mao, Negi, Narayan, Wang, Yang, Wang, Marcus, Khani~Shirkoohi, He, Nathan et~al.}]{park}
Mao, H.; Negi, P.; Narayan, A.; Wang, H.; Yang, J.; Wang, H.; Marcus, R.; Khani~Shirkoohi, M.; He, S.; Nathan, V.; et~al. 2019.
\newblock Park: An open platform for learning-augmented computer systems.
\newblock \emph{Advances in Neural Information Processing Systems}, 32.

\bibitem[{Mao, Netravali, and Alizadeh(2017)}]{pensieve}
Mao, H.; Netravali, R.; and Alizadeh, M. 2017.
\newblock Neural adaptive video streaming with pensieve.
\newblock In \emph{Proceedings of the Conference of the ACM Special Interest Group on Data Communication}, 197--210.

\bibitem[{Mao et~al.(2018)Mao, Venkatakrishnan, Schwarzkopf, and Alizadeh}]{mao2018variance}
Mao, H.; Venkatakrishnan, S.~B.; Schwarzkopf, M.; and Alizadeh, M. 2018.
\newblock Variance reduction for reinforcement learning in input-driven environments.
\newblock \emph{arXiv preprint arXiv:1807.02264}.

\bibitem[{Meng et~al.(2020)Meng, Wang, Bai, Xu, Mao, and Hu}]{metis}
Meng, Z.; Wang, M.; Bai, J.; Xu, M.; Mao, H.; and Hu, H. 2020.
\newblock Interpreting deep learning-based networking systems.
\newblock In \emph{Proceedings of the Annual conference of the ACM Special Interest Group on Data Communication on the applications, technologies, architectures, and protocols for computer communication}, 154--171.

\bibitem[{Miller(2019)}]{miller2019explanation}
Miller, T. 2019.
\newblock Explanation in artificial intelligence: Insights from the social sciences.
\newblock \emph{Artificial intelligence}, 267: 1--38.

\bibitem[{Mittelstadt, Russell, and Wachter(2019)}]{mittelstadt2019explaining}
Mittelstadt, B.; Russell, C.; and Wachter, S. 2019.
\newblock Explaining explanations in AI.
\newblock In \emph{Proceedings of the conference on fairness, accountability, and transparency}, 279--288.

\bibitem[{Mnih et~al.(2013)Mnih, Kavukcuoglu, Silver, Graves, Antonoglou, Wierstra, and Riedmiller}]{mnih2013playing}
Mnih, V.; Kavukcuoglu, K.; Silver, D.; Graves, A.; Antonoglou, I.; Wierstra, D.; and Riedmiller, M. 2013.
\newblock Playing atari with deep reinforcement learning.
\newblock \emph{arXiv preprint arXiv:1312.5602}.

\bibitem[{Mok, Chan, and Chang(2011)}]{mok2011measuring}
Mok, R.~K.; Chan, E.~W.; and Chang, R.~K. 2011.
\newblock Measuring the quality of experience of HTTP video streaming.
\newblock In \emph{12th IFIP/IEEE International Symposium on Integrated Network Management (IM 2011) and Workshops}, 485--492. IEEE.

\bibitem[{Patel, Abdu~Jyothi, and Narodytska(2024)}]{patel2024crystalbox}
Patel, S.; Abdu~Jyothi, S.; and Narodytska, N. 2024.
\newblock CrystalBox: Future-Based Explanations for Input-Driven Deep RL Systems.
\newblock \emph{Proceedings of the AAAI Conference on Artificial Intelligence}, 38(13): 14563--14571.

\bibitem[{Patel et~al.(2023)Patel, Zhang, Jyothi, and Narodytska}]{patel2023plume}
Patel, S.; Zhang, J.; Jyothi, S.~A.; and Narodytska, N. 2023.
\newblock Plume: A Framework for High Performance Deep RL Network Controllers via Prioritized Trace Sampling.
\newblock arXiv:2302.12403.

\bibitem[{Pohlen et~al.(2018)Pohlen, Piot, Hester, Azar, Horgan, Budden, Barth-Maron, Van~Hasselt, Quan, Ve{\v{c}}er{\'\i}k et~al.}]{pohlen2018observe}
Pohlen, T.; Piot, B.; Hester, T.; Azar, M.~G.; Horgan, D.; Budden, D.; Barth-Maron, G.; Van~Hasselt, H.; Quan, J.; Ve{\v{c}}er{\'\i}k, M.; et~al. 2018.
\newblock Observe and look further: Achieving consistent performance on atari.
\newblock \emph{arXiv preprint arXiv:1805.11593}.

\bibitem[{Puri et~al.(2019)Puri, Verma, Gupta, Kayastha, Deshmukh, Krishnamurthy, and Singh}]{puri2019explain}
Puri, N.; Verma, S.; Gupta, P.; Kayastha, D.; Deshmukh, S.; Krishnamurthy, B.; and Singh, S. 2019.
\newblock Explain your move: Understanding agent actions using specific and relevant feature attribution.
\newblock \emph{arXiv preprint arXiv:1912.12191}.

\bibitem[{Ribeiro, Singh, and Guestrin(2016)}]{ribeiro2016should}
Ribeiro, M.~T.; Singh, S.; and Guestrin, C. 2016.
\newblock " Why should i trust you?" Explaining the predictions of any classifier.
\newblock In \emph{Proceedings of the 22nd ACM SIGKDD international conference on knowledge discovery and data mining}, 1135--1144.

\bibitem[{Rotman, Schapira, and Tamar(2020)}]{rotman2020online}
Rotman, N.~H.; Schapira, M.; and Tamar, A. 2020.
\newblock Online safety assurance for learning-augmented systems.
\newblock In \emph{Proceedings of the 19th ACM Workshop on Hot Topics in Networks}, 88--95.

\bibitem[{Silver(2015)}]{silver2015}
Silver, D. 2015.
\newblock Lectures on Reinforcement Learning.
\newblock \url {https://www.davidsilver.uk/teaching/}.
\newblock Accessed: 2022-10-12.

\bibitem[{Sutton and Barto(2018)}]{sutton2018reinforcement}
Sutton, R.~S.; and Barto, A.~G. 2018.
\newblock \emph{Reinforcement learning: An introduction}.
\newblock MIT press.

\bibitem[{van~der Waa et~al.(2018)van~der Waa, van Diggelen, Bosch, and Neerincx}]{van2018contrastive}
van~der Waa, J.; van Diggelen, J.; Bosch, K. v.~d.; and Neerincx, M. 2018.
\newblock Contrastive explanations for reinforcement learning in terms of expected consequences.
\newblock \emph{arXiv preprint arXiv:1807.08706}.

\bibitem[{Verma et~al.(2018)Verma, Murali, Singh, Kohli, and Chaudhuri}]{verma2018programmatically}
Verma, A.; Murali, V.; Singh, R.; Kohli, P.; and Chaudhuri, S. 2018.
\newblock Programmatically interpretable reinforcement learning.
\newblock In \emph{International Conference on Machine Learning}, 5045--5054. PMLR.

\bibitem[{Yan et~al.(2020)Yan, Ayers, Zhu, Fouladi, Hong, Zhang, Levis, and Winstein}]{fugu}
Yan, F.~Y.; Ayers, H.; Zhu, C.; Fouladi, S.; Hong, J.; Zhang, K.; Levis, P.; and Winstein, K. 2020.
\newblock Learning in situ: a randomized experiment in video streaming.
\newblock In \emph{17th USENIX Symposium on Networked Systems Design and Implementation (NSDI 20)}, 495--511.

\bibitem[{Yau, Russell, and Hadfield(2020)}]{yau2020did}
Yau, H.; Russell, C.; and Hadfield, S. 2020.
\newblock What did you think would happen? explaining agent behaviour through intended outcomes.
\newblock \emph{Advances in Neural Information Processing Systems}, 33: 18375--18386.

\bibitem[{Zahavy, Ben-Zrihem, and Mannor(2016)}]{zahavy2016graying}
Zahavy, T.; Ben-Zrihem, N.; and Mannor, S. 2016.
\newblock Graying the black box: Understanding dqns.
\newblock In \emph{International conference on machine learning}, 1899--1908. PMLR.

\bibitem[{Zhang, Zhou, and Lin(2020)}]{zhang2020interpretable}
Zhang, H.; Zhou, A.; and Lin, X. 2020.
\newblock Interpretable policy derivation for reinforcement learning based on evolutionary feature synthesis.
\newblock \emph{Complex \& Intelligent Systems}, 6(3): 741--753.

\end{thebibliography}

\appendix

\begin{figure*}[t]
\centering
\begin{subfigure}{.99\textwidth}
    \centering
    \includegraphics[width=0.6\textwidth]{mse_legend.png}
\end{subfigure}
\newline
\begin{subfigure}[t]{.65\textwidth}
    \centering
    \includegraphics[width=\textwidth]{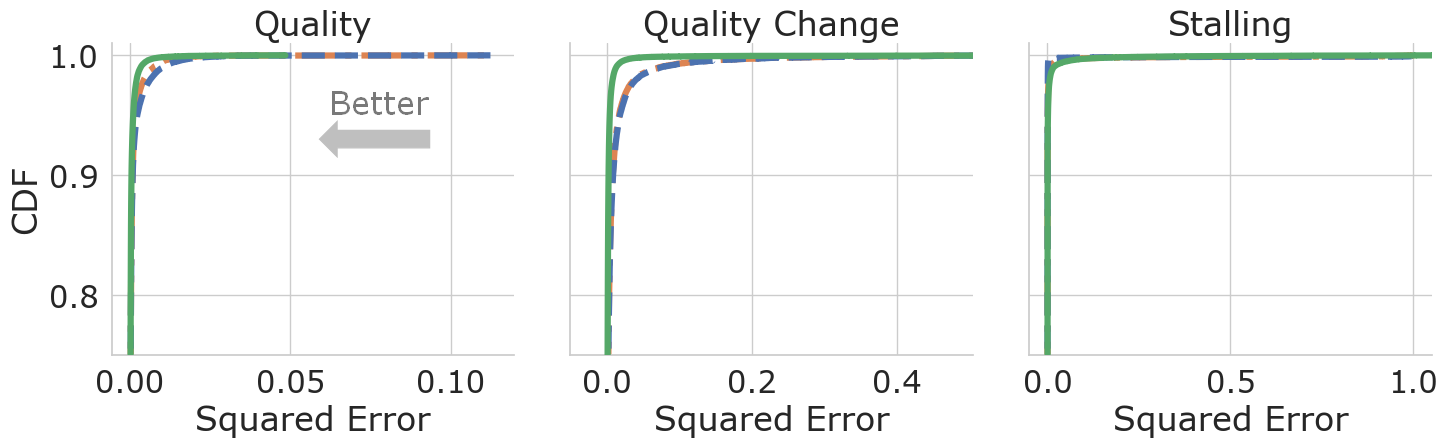}
    \caption{\small (a) Factual Actions}
\end{subfigure}
\begin{subfigure}[t]{.65\textwidth}
    \centering
    \includegraphics[width=\textwidth]{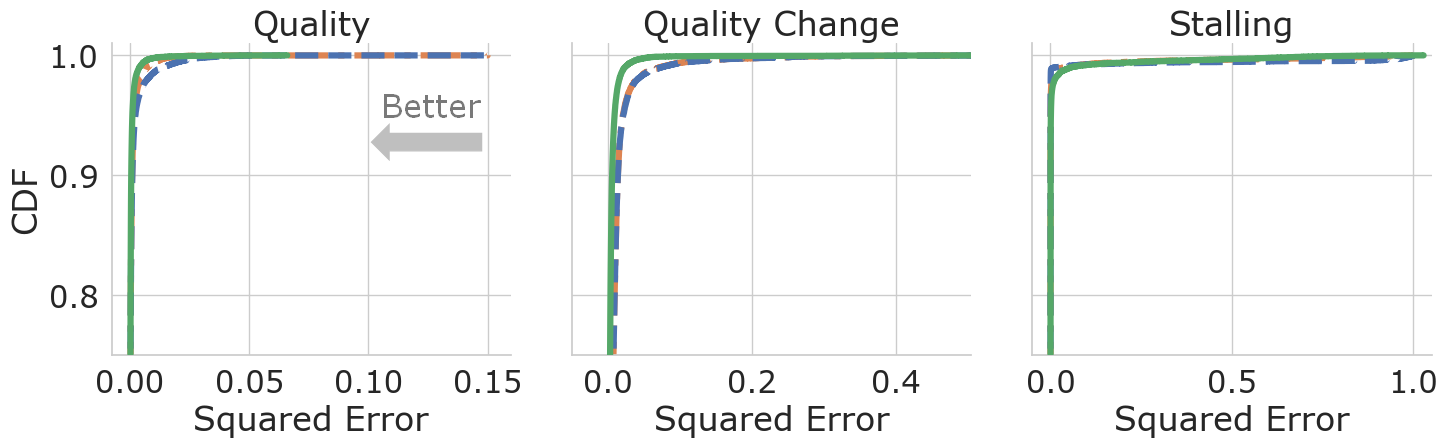}
    \caption{\small (b) Counter Factual Actions}
\end{subfigure}

\caption{\small \textbf{Evaluation of CrystalBox in ABR across all traces}. Distribution of Squared Error to samples of the ground truth decomposed return predictions for all traces in ABR. We observe that the differences in the distribution of error for all of the return predictors shrink, but the relative ordering remains the same: CrystalBox offers high-fidelity predictions for both factual and counterfactual actions.}
\label{fig:abr_all_traces_mse}
\end{figure*}

\begin{figure*}[t]
\centering
\begin{subfigure}{.99\textwidth}
    \centering
    \includegraphics[width=0.475\textwidth]{event_detection_legend.png}
\end{subfigure}
\newline
\begin{subfigure}[t]{.4\textwidth}
    \centering
    \includegraphics[width=0.89\textwidth]{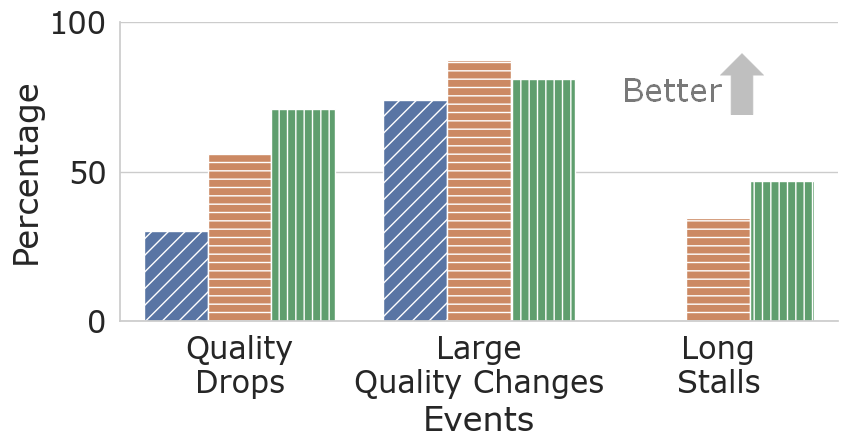}
    \caption{\small (a) ABR: Recall}
\end{subfigure}
\begin{subfigure}[t]{.4\textwidth}
    \centering
    \includegraphics[width=0.899\textwidth]{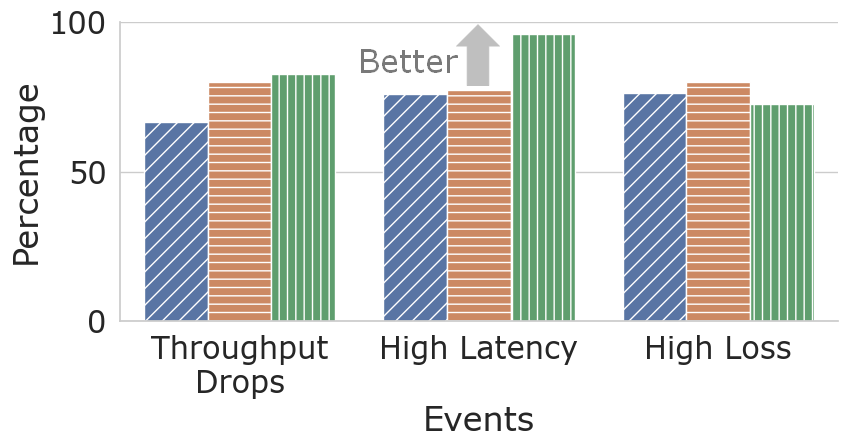}
    \caption{\small (b) CC: Recall}
\end{subfigure}
\begin{subfigure}[t]{.4\textwidth}
    \centering
    \includegraphics[width=0.89\textwidth]{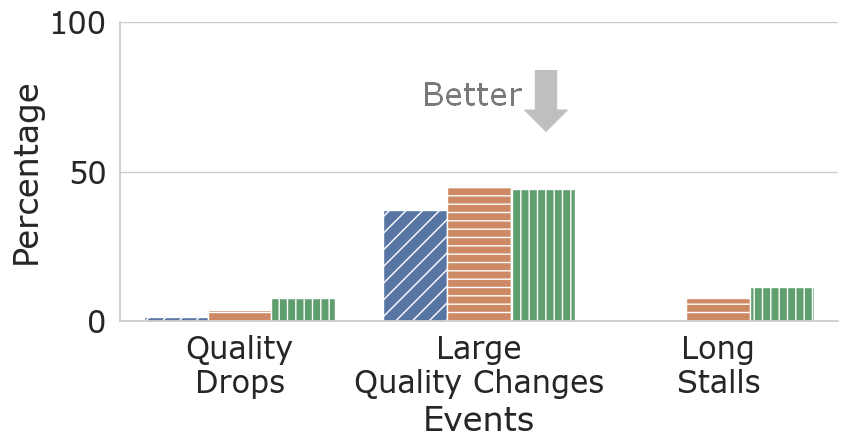}
    \caption{\small (c) ABR: False Positive Rate}
\end{subfigure}
\begin{subfigure}[t]{.4\textwidth}
    \centering
    \includegraphics[width=0.89\textwidth]{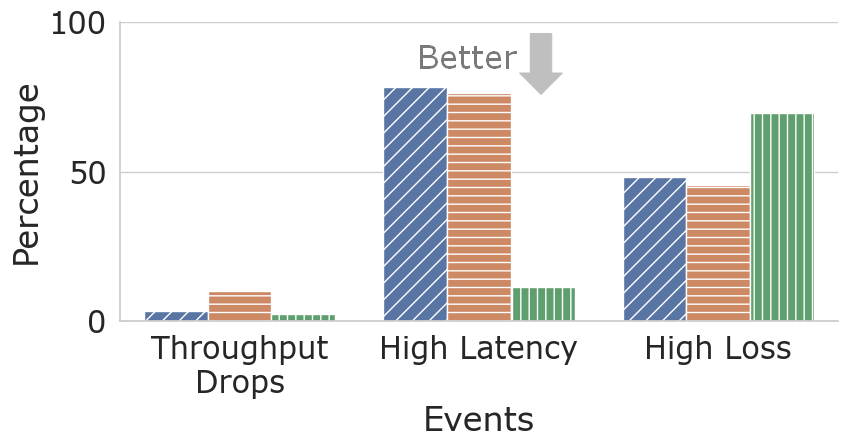}
    \caption{\small (d) CC: False Positive Rate}
\end{subfigure}
\caption{\small \textbf{Large Performance Drop Event Detection under counter-factual actions}: We analyze the efficacy of different predictors for detecting large performance drops. We identify events happening by detecting if samples of the ground-truth return exceed a threshold. We evaluate their efficacy by analyzing both their recall and their false positive rates under counterfactual actions.}
\label{fig:event_detection_counter_factual}
\end{figure*}

\begin{figure*}[!t]
\centering
\begin{subfigure}[t]{.8\textwidth}
    \centering
    \includegraphics[width=0.99\textwidth]{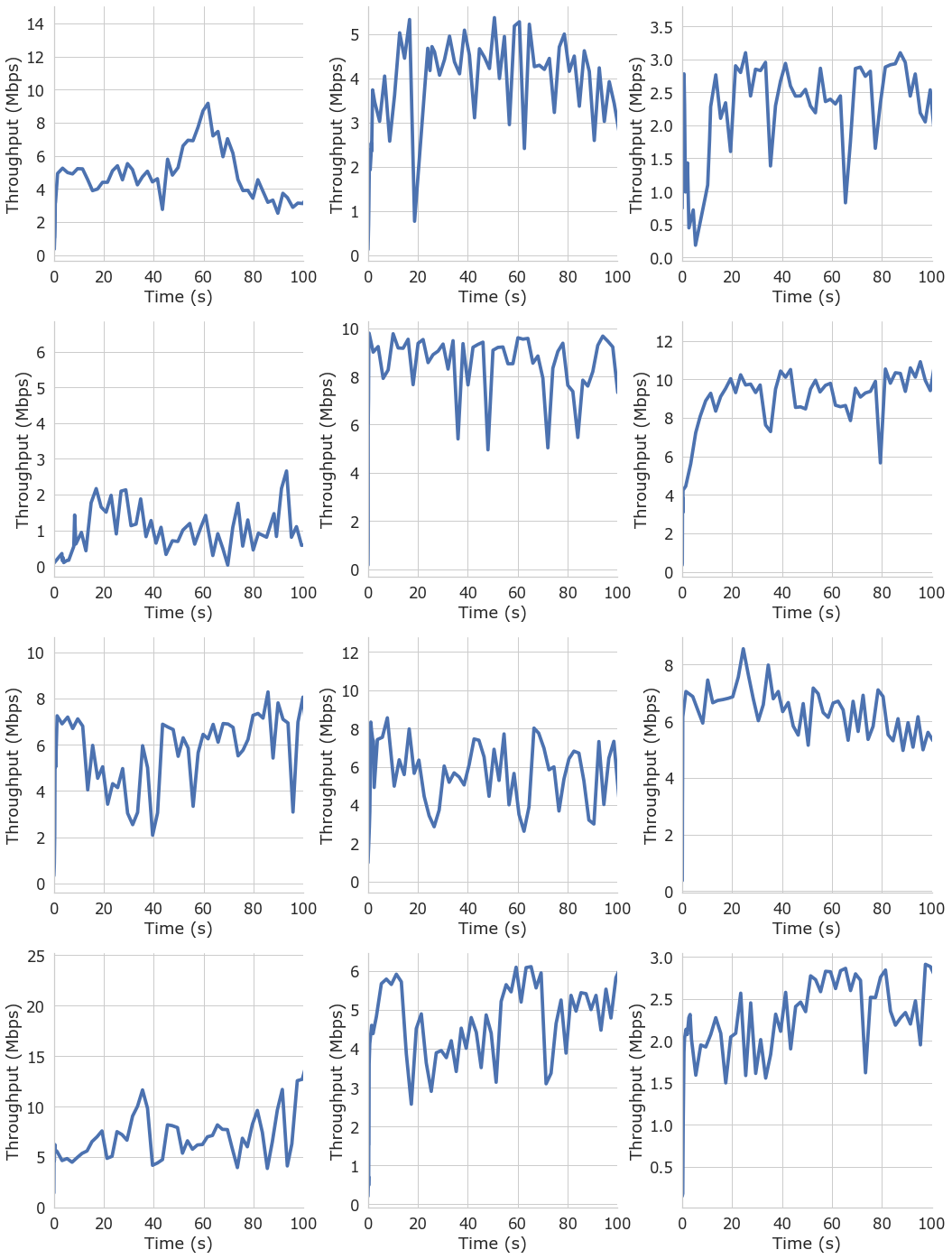}
\end{subfigure}
\caption{\small \textbf{Examples of Traces in Adaptive Bitrate Streaming}. In ABR, a trace is the over-time throughput of the internet connection between a viewer and a streaming platform. In this figure, we present a visualization of a few of those traces for the first 100 seconds. Note that the y-axis is different on each plot due to inherent differences between traces.}
\label{fig:abr_trace_example}
\vspace{10mm}
\end{figure*}

\begin{figure*}[!t]
\centering
\begin{subfigure}[t]{.8\textwidth}
    \centering
    \includegraphics[width=0.99\textwidth]{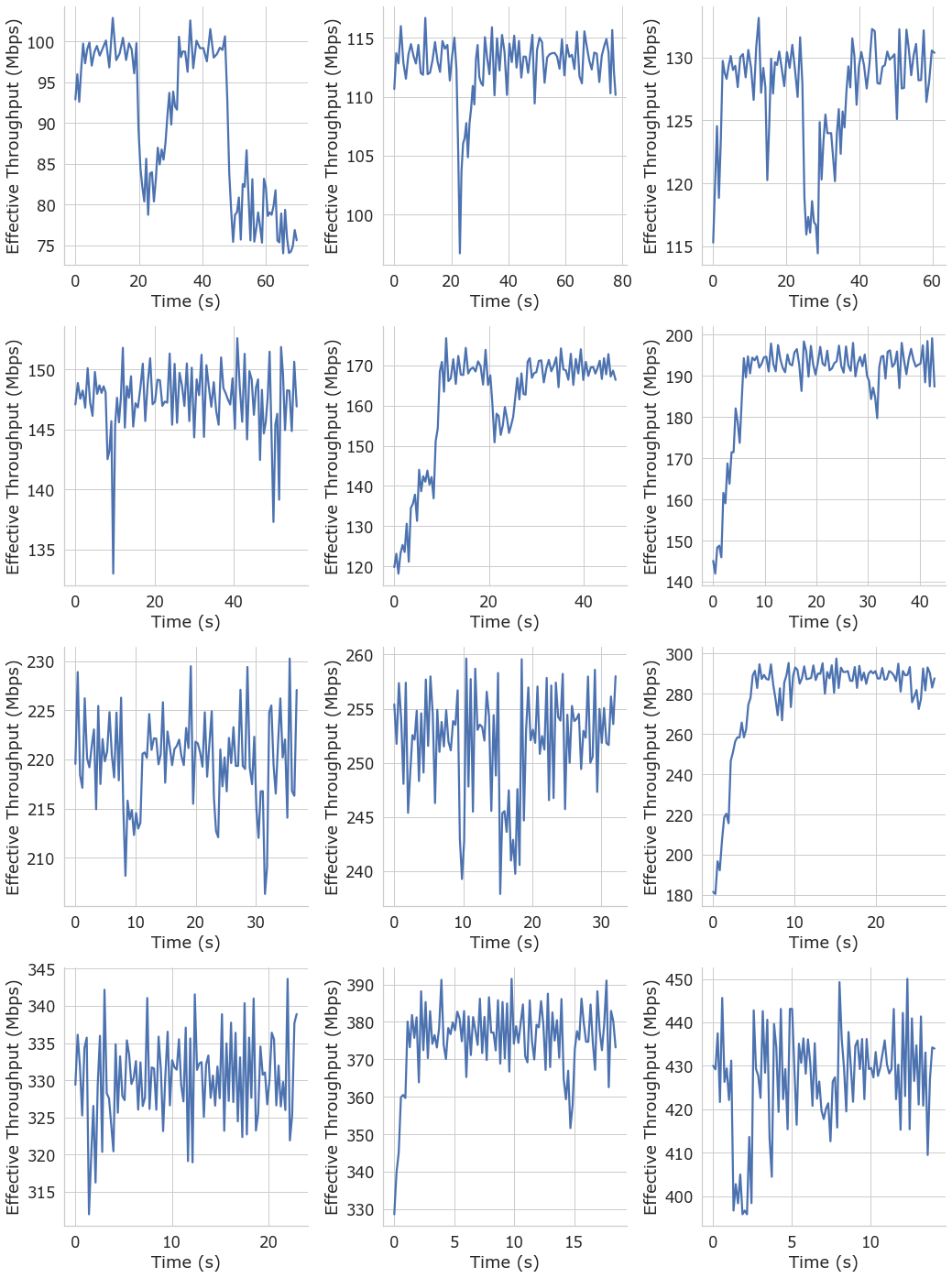}
\end{subfigure}
\caption{\small \textbf{Examples of Traces in Congestion Control}. In CC, a trace is defined as the internet network conditions between a sender and a receiver over time. These conditions can be characterized by many different metrics such as throughput, latency, or loss. In this figure, we represent these traces by the sender's effective throughput over time. Note that the x-axis and y-axis differ on each plot due to the inherent differences between traces. }
\label{fig:cc_trace_example}
\end{figure*}

\begin{figure*}[!t]
\centering
\begin{subfigure}[t]{.625\textwidth}
    \centering
    \includegraphics[width=0.99\textwidth]{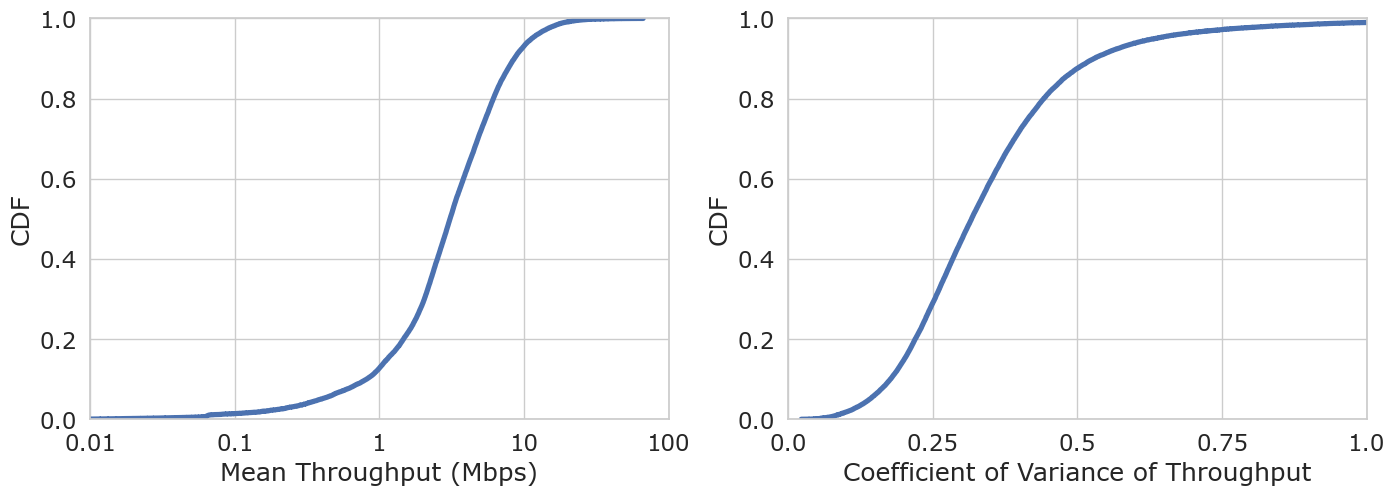}
\end{subfigure}
\begin{subfigure}[t]{.26\textwidth}
    \centering
    \includegraphics[width=0.99\textwidth]{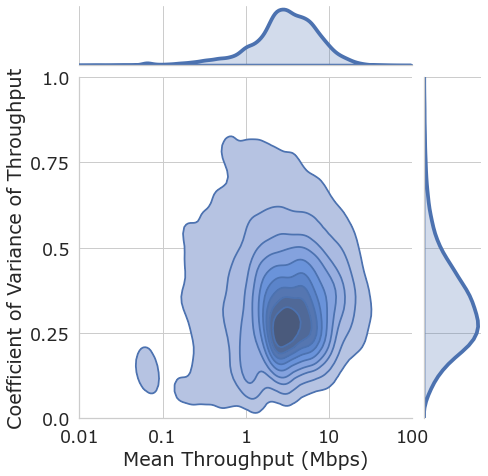}
\end{subfigure}
\caption{\small Distribution of Traces in ABR. \textbf{Left}: distribution of the mean throughput in traces. Note that the x-axis is log-scaled due to the large differences between all the clients of this server. \textbf{Middle}: distribution of coefficient of variance of the throughput within each trace. \textbf{Right}: The joint distribution of mean and coefficient of variance of throughput. The traces are logged over the course of a couple of months from an online public live-streaming Puffer~\cite{fugu}. We find that a majority of the traces have mean throughput well above the bitrate of the highest quality video. Only a small percentage of traces represent poor network conditions such as low throughput, high variance, etc.}
\label{fig:abr_trace_stats}
\end{figure*}

\begin{figure*}[!t]
\centering
\begin{subfigure}[t]{.999\textwidth}
    \centering
    \includegraphics[width=0.999\textwidth]{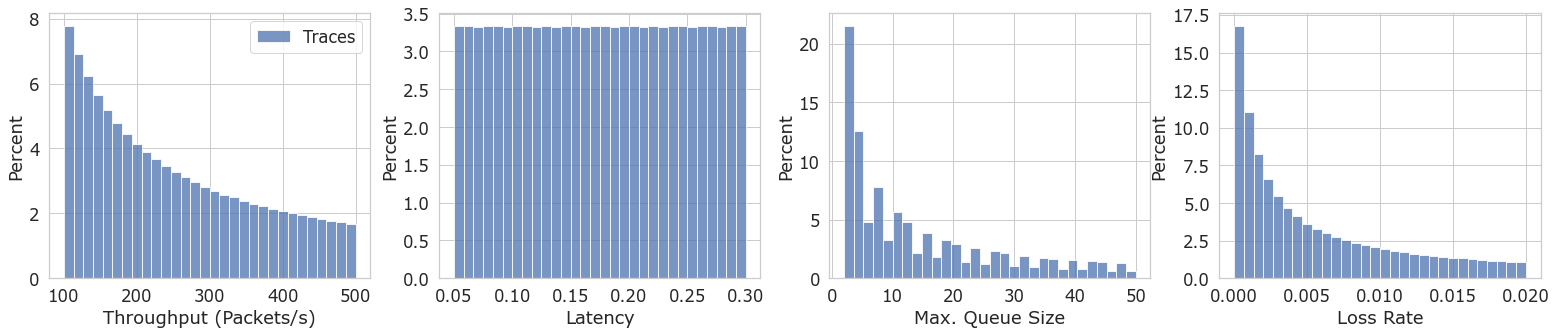}
\end{subfigure}
\caption{\small \textbf{Distribution of Traces in CC}: We analyze the distribution of traces in CC by analyzing the distribution by four key metrics: throughput, latency, maximum queue size, and loss. These traces are synthetically generated by sampling from a range of values, similar to the technique employed by \cite{aurora}. We observe that traces with especially poor network conditions such as high loss rate or high queuing delay are small in number.}
\label{fig:cc_trace_stats}
\end{figure*}

\begin{figure*}[t]
\centering
\begin{subfigure}[t]{.6\textwidth}
    \centering
    \includegraphics[width=0.99\textwidth]{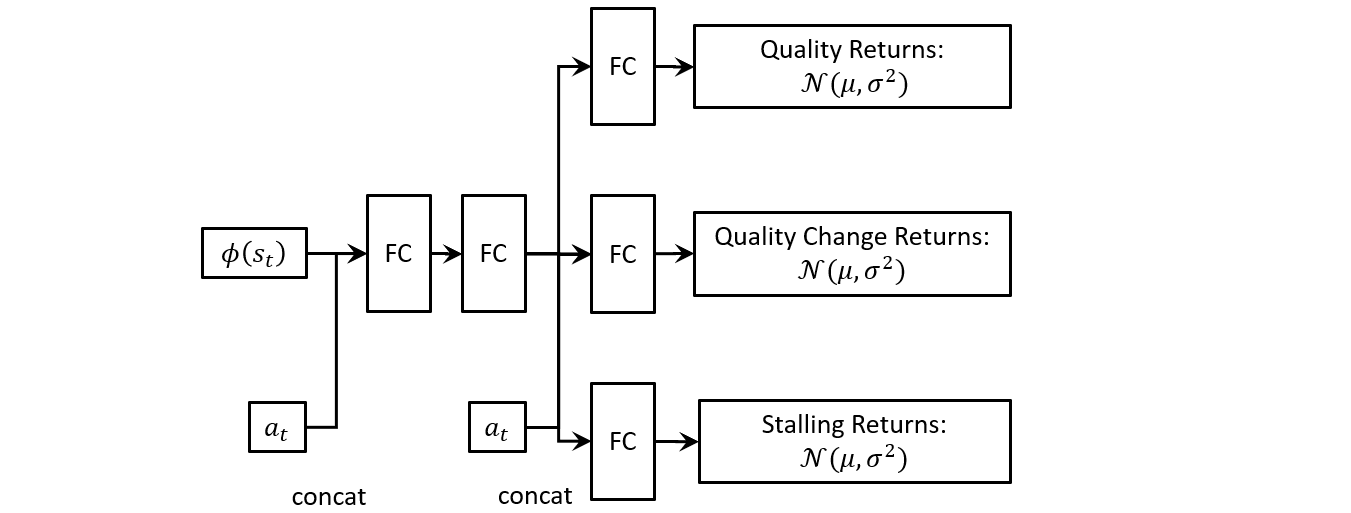}
\end{subfigure}
\caption{\small \textbf{Neural Architecture of CrystalBox's Learned Predictor in ABR}.}
\label{fig:abr_model}
\end{figure*}

\begin{figure*}[t]
\centering
\begin{subfigure}[t]{.6\textwidth}
    \centering
    \includegraphics[width=0.99\textwidth]{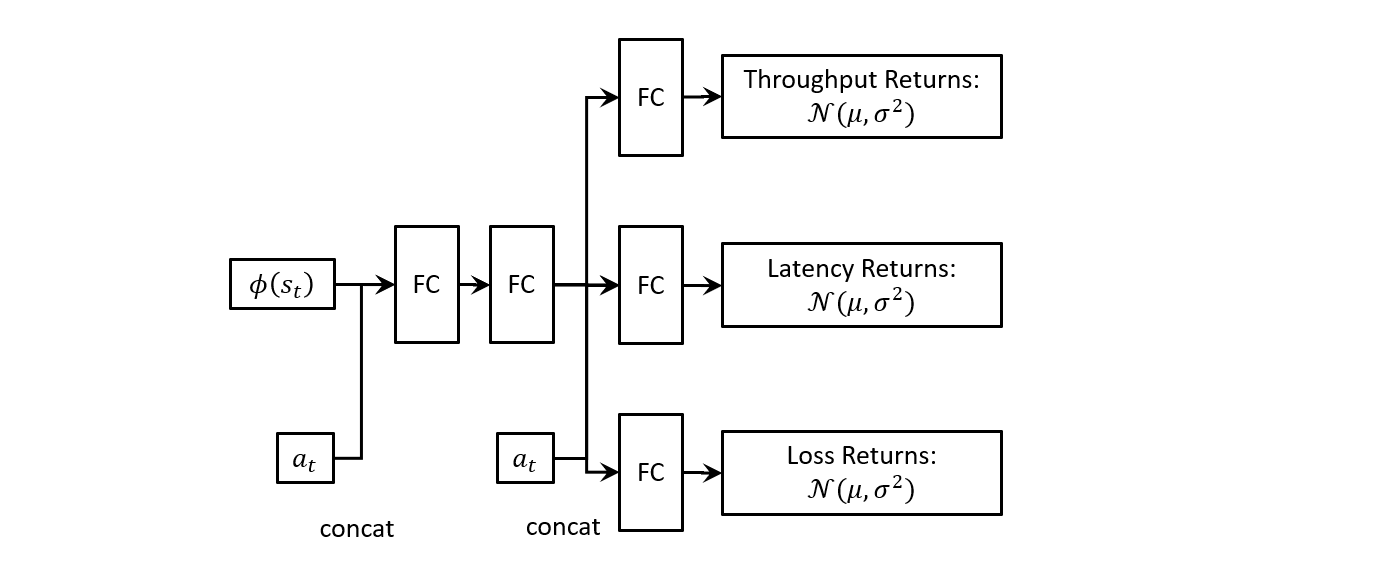}
\end{subfigure}
\caption{\small \textbf{Neural Architecture of CrystalBox's Learned Predictor in CC}.}
\label{fig:cc_model}
\end{figure*}

\clearpage
\section{Appendix}
\subsection{Traces}
\label{subsec:traces}

In this section, we visualize representative traces in Figure~\ref{fig:abr_trace_example} and Figure~\ref{fig:cc_trace_example} for ABR and CC applications, respectively.  

In ABR, a trace is the over-time throughput of the internet connection between a viewer and a streaming platform. We obtain a representative set of traces by analyzing the logged data of a public live-streaming platform~\cite{fugu}. In Figure~\ref{fig:abr_trace_example} we present a visualization of a few of those traces for the first 100 seconds. Note that the y-axis is different on each plot due to inherent differences between traces. However, even with the naked eye, we can see that some traces are high-throughput traces, e.g. all traces in the third row, while other traces are slow-throughput, e.g. the first plot in the second row. To further analyze these inherent differences, we analyze the distribution of mean and coefficient of variance of throughput within each trace. In Figure~\ref{fig:abr_trace_stats}, we see that a majority of traces have high mean throughput. When we analyze this jointly with the distribution of the throughput coefficient of variance, we see that a majority of those traces also have smaller variances. Only a small number of traces represent poor network conditions such as low throughput or high throughput variance. These observations are consistent with a recent Google study~\cite{langley2017quic} that showed that more than $93\%$ of YouTube streams \textit{never} come to a stall.     

In CC, a trace is the over-time network conditions between a sender and receiver. We obtain a representative set of traces by following \citep{aurora} and synthetically generating them by four key values: mean throughput, latency, queue size, and loss rate. In Figure~\ref{fig:cc_trace_example}, we demonstrate how these traces may look like from the sender's perspective by looking at the effective throughput over time. Similar to the traces in ABR, we can visually see that the traces can be greatly different from one another. In Figure~\ref{fig:cc_trace_stats}, we analyze the effective distribution of these traces. We observe that while the distribution is not nearly as unbalanced as it is in ABR, there are still only a small number of traces that have exceedingly harsh network conditions.

\subsection{Network Observability (Counterfactual actions)}
\label{subsec:event_detect_appendix}
We present the performance of CrystalBox for network observability by analyzing its ability to rise alerts about upcoming large performance drops under counterfactual actions. We recall that our goal is to detect states and actions that lead to large performance drops. We employ CrystalBox's optional post-processing and convert vectors of output values into binary events. In our experiments, we used the following thresholds. For ABR, we used: quality return below 0.55, quality change return below -0.1, and stalling return below -0.25. For CC, we used: throughput return below 0.3, latency return below -0.075, and loss return below -0.1.

In Section~\ref{subsec:event_detect_eval}, we analyzed the performance of CrystalBox's ability to detect these events under factual actions (actions that the policy takes). Now, we turn to present the results of the same state under comparative counterfactual actions. In Figure~\ref{fig:event_detection_counter_factual}, we present the recall and false-positive rates of different return predictors. Similar to the results under factual actions, we find that CrystalBox has higher recall and lower false-positive rates. In ABR, we see that CrystalBox achieves significantly higher recall in detecting quality drop and stalling events while having about $5\%$ higher false-positive rates. We additionally see that Distribution-Aware sampling achieves significantly higher recall than Naive sampling, particularly in long stall events. In CC, we see that CrystalBox is particularly adept at detecting throughput drop and high latency events, but suffers from high false-positive rates of high loss events. 

\subsection{Fidelity Evaluation (additional results)}
\label{subsec:additional_fidelity}

We present our evaluation of \fmname explanations on all traces. Figure~\ref{fig:abr_all_traces_mse} shows our results. We can see that all predictors perform well. For high throughput traces, the optimal policy for the controller is simple: send the highest bitrate. Therefore, all predictors do well on these traces. However, the relative performance between the predictors is the same as it was with traces that could experience stalling and quality drops~\ref{fig:main_mse_results}. The CrystalBox outperforms sampling-based methods across all three reward components under both factual and counterfactual actions. 

\subsection{Policy Rollouts}
We collect samples of the ground truth values of the decomposed future returns by rolling out the policy in a simulation environment. That is, we let the policy interact with the environment under an offline set of traces $Z$, and observe sequences of the tuple $(s, a, \vec{r})$. With these tuples, we can calculate the decomposed return $Q_c^\pi(s_t, a_t)$ for each timestamp. In practice, we do not use an infinite sum to obtain $Q^\pi_c$. This is because the variance of this function can be close to infinity due to its dependence on the length of $z$, which can range from 1 to infinity~\cite{mao2018variance}. Instead, we bound this sum by a constant $t_{max}$. This approximates $Q^\pi_c$ with a commonly used truncated version where the rewards after $t_{max}$ are effectively zero~\cite{sutton2018reinforcement}. 

For a given episode, these states and returns can be highly correlated~\cite{mnih2013playing}. Thus, to efficiently cover a wide variety of scenarios, we do not consider the states and returns $Q_c^\pi(s_t, a_t)$ after $s_t$ for $t_{max}$ steps. Moreover, when attempting to collect samples for a counterfactual action $a_t'$, we ensure the rewards and actions from timestamp $t$ onwards are not used in the calculation for any state-action pair before $(s_t, a_t')$. This strategy avoids adding any additional noise to samples of $Q^\pi$ due to exploratory actions.

$t_{max}$ is a hyper-parameter for each environment. 
In systems environments, we usually observe the effect of each action within a short time horizon. For example, if a controller drops bitrate, then the user experiences lower quality video in one step. Therefore, it is only required to consider rollouts of a few steps to capture the consequences of each action, so $t_{max}$ of five is sufficient for our environments.

\subsection{CrystalBox Details}
\label{subsec:appendix_crystalbox_details}
\parab{Preprocessing}. We employ Monte Carlo Rollouts to get samples of $Q^\pi_c$ for training our learned predictor. By themselves, the return components can vary across multiple orders of magnitude. Thus, similar to the standard reward clipping~\citep{mnih2013playing} and return normalization~\citep{pohlen2018observe} techniques widely employed in Q-learning, we normalize all the returns to be in the range [0, 1].

\parab{Neural Architecture Design}.
We design the neural architecture of our learned predictors to be compact and sample efficient. We employ shared layers that feed into separate fully connected `tails' that then predict the return components. We model the samples of $Q_c^{\pi}$ as samples from a Gaussian distribution and predict the parameters (mean and standard deviation) to this distribution in each tail. To learn to predict these parameters, we minimize the negative log-likelihood loss of each sample of $Q_c^{\pi}$. 

For the fully connected layers, we perform limited tuning to choose the units of these layers from \{64, 128, 256, 512\}. We found that a smaller number of units is enough in both of our environments. We present a visualization of our architectures in Figures~\ref{fig:abr_model} and \ref{fig:cc_model}. 

\parab{Learning Parameters}.
We learn our predictors in two stages. In the first stage, we train our network end-to-end. In the second stage, we freeze the shared weights in our network and fine-tune our predictors with a smaller learning rate. We use an Adagrad optimizer, and experimented with learning rates from 1e-6 to 1e-4, with decay from 1e-10 to 1e-9. We tried batch sizes from \{50, 64, 128, 256, 512\}. We found that small batch sizes, learning rates, and decay work best.

\end{document}